\documentclass[10pt]{article}
\usepackage{graphicx} 
\PassOptionsToPackage{sort&compress,numbers,super}{natbib}
\usepackage[preprint]{neurips_2022}

\usepackage[utf8]{inputenc}
\usepackage[T1]{fontenc}   
\usepackage[hidelinks]{hyperref}    
\usepackage{url}            
\usepackage{booktabs}    
\usepackage{amsfonts}       
\usepackage{nicefrac}       
\usepackage{microtype}      
\usepackage{xcolor}         
\usepackage{microtype}
\usepackage{caption}
\usepackage{tikz}
\usetikzlibrary{matrix, arrows}
\usepackage[width=474.18663pt]{caption}
\usepackage{subfigure}
\usepackage[most]{tcolorbox}
\usepackage{fvextra}
\usepackage{csquotes}
\usepackage{float}
\usepackage{alltt}
\usepackage{soul}
\usepackage{fancyvrb}
\usepackage{multirow}
\usepackage{tikz}
\usepackage[most]{tcolorbox}
\usepackage{xcolor}
\definecolor{block-quote}{gray}{0.95}
\newtcolorbox{myquote}{colback=block-quote,grow to right by=-10mm,grow to left by=-10mm,boxrule=0pt,boxsep=0pt,breakable}
\usetikzlibrary{shapes,calc,positioning}

\global

\usepackage{soul}

\tcbset{
  aibox/.style={
    width=474.18663pt,
    top=10pt,
    colback=white,
    colframe=black,
    colbacktitle=black,
    enhanced,
    center,
    attach boxed title to top left={yshift=-0.1in,xshift=0.15in},
    boxed title style={boxrule=0pt,colframe=white,},
  }
}
\newtcolorbox{AIbox}[2][]{aibox,title=#2,#1}

\definecolor{aigold}{RGB}{244,210, 1} 
\definecolor{aigreen}{RGB}{210,244,211} 

\sethlcolor{aigreen}

\definecolor{aired}{RGB}{255,180,181}

\newtcbox{\mybox}[1][green]{on line,
arc=0pt,outer arc=0pt,colback=#1!10!white,colframe=#1!50!black,
boxsep=0pt,left=0pt,right=0pt,top=0pt,bottom=0pt,
boxrule=0pt,bottomrule=0pt,toprule=0pt}

\title{Towards Large-scale Chemical Reaction Image Parsing via a Multimodal Large Language Model}

\author{
Yufan CHEN$^{1}$ \quad Ching Ting LEUNG$^{1}$  \quad  Jianwei SUN$^{1,3}$  \quad \\
\textbf{Yong HUANG}$^{3}$ \quad \textbf{Linyan LI}$^{4*}$ \quad \textbf{Hao CHEN}$^{1,2*}$  \quad  \textbf{Hanyu GAO}$^{1*}$  \\
$^1$ Department of Chemical and Biological Engineering, Hong Kong \\University of Science and Technology, Hong Kong SAR, China\\ 
$^2$ Department of Computer Science and Engineering, Hong Kong \\University of Science and Technology, Hong Kong SAR, China\\ 
$^3$ Department of Chemistry, Hong Kong University of Science and\\ Technology, Hong Kong SAR, China\\ 
$^4$ Department of Data Science, City University of Hong Kong,\\ Hong Kong SAR, China\\
$^*$ Corresponding author(s).\\
\texttt{ychenkv@connect.ust.hk}, \texttt{linyanli@cityu.edu.hk},\\
\texttt{jhc@cse.ust.hk}, \texttt{hanyugao@ust.hk}
}

\begin{document}
\begin{sloppypar}
\maketitle
\begin{abstract}
Artificial intelligence (AI) has demonstrated significant promise in advancing organic chemistry research; however, its effectiveness depends on the availability of high-quality chemical reaction data. Currently, most published chemical reactions are not available in machine-readable form, limiting the broader application of AI in this field. The extraction of published chemical reactions into structured databases still relies heavily on manual curation, and robust automatic parsing of chemical reaction images into machine-readable data remains a significant challenge. To address this, we introduce the Reaction Image Multimodal large language model (RxnIM), the first multimodal large language model specifically designed to parse chemical reaction images into machine-readable reaction data. RxnIM not only extracts key chemical components from reaction images but also interprets the textual content that describes reaction conditions. Together with specially designed large-scale dataset generation method to support model training, our approach achieves excellent performance, with an average F1 score of 88\% on various benchmarks, surpassing literature methods by 5\%. This represents a crucial step toward the automatic construction of large databases of machine-readable reaction data parsed from images in the chemistry literature, providing essential data resources for AI research in chemistry. The source code, model checkpoints, and datasets developed in this work are released under permissive licenses. An instance of the RxnIM web application can be accessed at \url{https://huggingface.co/spaces/CYF200127/RxnIM}.
\end{abstract}

\section{Introduction}
\label{sec:introduction}
%In the field of synthetic chemistry, numerous chemical reaction schemes represented by images are presented in the literature.
%As shown in Fig.~\ref{fig:img_example}, these images contain essential information in the reaction schemes, such as molecular structures or condition texts, and can be extremely complex.
The field of organic chemistry has witnessed a transformative shift with machine learning techniques, enabling significant advancements in retrosynthesis, reaction prediction, and condition recommendation.As researchers increasingly leverage these methodologies to explore complex chemical phenomena, high-quality machine-readable chemical reaction data is essential. Despite the wealth of chemical knowledge documented in the literature,the data required for effective machine learning applications remains largely fragmented and predominantly inaccessible in a format suitable for computational analysis.~\cite{1staker2019molecular,2beard2020chemschematicresolver,3wilary2021reactiondataextractor,4wilary2023reactiondataextractor,5qian2023rxnscribe}.

Traditional approaches to chemical reaction data extraction are predominantly manual, involving labor-intensive curation processes that are susceptible to human error and inefficiencies. This reliance on manual extraction limits the scalability of data acquisition as well as the potential for comprehensive analysis across large datasets. There is an urgent need for automated solutions that can accurately and efficiently parse chemical reaction images, transforming them into structured data that can support advanced machine learning applications.

Substantial efforts have been devoted to automatic chemical reaction data extraction.~\cite{11jessop2011oscar4,12hawizy2011chemicaltagger,13lowe2012extraction,14swain2016chemdataextractor,18steiner2019organic,19vaucher2020automated}. 
 For instance, the Pistachio dataset~\cite{15mayfield2018pistachio}, primarily derived from patent text, utilizes a classic natural language processing pipeline that encompasses syntactic parsing and named entity recognition to identify chemical names~\cite{16lowe2020extraction,17nguyen2020chemu}, followed by event extraction to organize these chemicals into reactions. 
 To handle the varied text found in journal articles, Guo et al. developed a deep learning approach~\cite{20guo2021automated} that breaks down the task into product extraction and reaction role labeling, utilizing sequence tagging techniques based on pre-trained language models. 
 Ming et al. proposed a reaction extraction system based on large language models (LLMs)~\cite{21zhong2023reaction}, which utilized the natural language understanding ability of LLMs, and expanded the scope of existing predefined reaction roles to include important attributes that have been ignored before, thus providing a more comprehensive and accurate description of chemical reactions. Despite these advancements, existing work primarily focus only on processing textual information. 
 
 Images serve as a more intuitive medium for documenting chemical reactions, providing clear visualization of molecular structures and the logical flow of multi-component and multi-step reactions. Yet, reaction image parsing remains under-explored, largely due to the complexity of reaction images and the variability of drawing styles. Previous works have attempted to detect the location of molecular objects from images~\cite{3wilary2021reactiondataextractor,4wilary2023reactiondataextractor} or recognizing their chemical structures~\cite{qian2023molscribe,22filippov2009optical,23rajan2020decimer,24oldenhof2020chemgrapher}. These methods often struggle to understand the role of different components and their logical connections, which are critical for a comprehensive understanding of the chemical reactions depicted~\cite{3wilary2021reactiondataextractor,4wilary2023reactiondataextractor}. 
For reaction image parsing, Qian et al. developed a single encoder-decoder model~\cite{5qian2023rxnscribe} closely following Pix2Seq~\cite{chen2021pix2seq} attempting to parse reaction data from images directly via an image-to-sequence translation. This approach demonstrated strong promise, yet it still frequently failed to parse data from images of more complicated reaction patterns.
Additionally, the multimodality of reaction information has not been explicitly addressed in previous methods. For example, for the text of reaction conditions and other auxiliary information, existing methods typically rely on external optical character recognition (OCR) tools to recognize the characters, and do not further process the information (e.g., whether the text describes the reagents, solvents, time, temperature, or yield), resulting in less comprehensive final parsed data.

Recently, as an essential subset of LLMs~\cite{openai2022chatgpt,31brown2020languageGPT3,ouyang2022trainingINSGPT,raffel2020exploringT5,chowdhery2023palm,zhang2022opt}, multimodal large language models (MLLMs) represent a significant breakthrough in computer vision~\cite{alayrac2022flamingo,shen2024hugginggpt,yang2023mmGPT,wu2023visualGPT,li2023videochat,liu2023internchat}. They have demonstrated impressive capabilities in both traditional visual tasks, such as object detection and instance segmentation, as well as in more complex tasks like referring expression comprehension and  generation~\cite{liu2023hidden}. Moreover, MLLMs have shown exceptional OCR capabilities across various text-related visual tasks, even without specialized training on relevant datasets. Therefore, MLLMs present a promising solution for parsing data from chemical reaction images, an area that remains largely uncharted in the literature. 

In this paper, we present Reaction Image Multimodal large language model (RxnIM), the first MLLM specifically designed for chemical reaction image parsing. The model was trained using a three-stage training strategy with a unified language-based task instruction for different chemical reaction image parsing tasks. The first stage was pretraining the model's object detection capability on a large dataset of synthetic reaction images. In the second stage, the model was trained to identify the reaction components and extract reaction conditions using the synthetic dataset. In the final stage, the model was fine-tuned on a smaller, manually curated dataset to enhance its performance on real reaction images. Unlike previous approaches that predict object types and relationships based on predefined rules~\cite{3wilary2021reactiondataextractor,4wilary2023reactiondataextractor} or single-task models~\cite{5qian2023rxnscribe}, RxnIM comprehends the entire reaction process depicted in the reaction image and integrates various parsing tasks seamlessly and flexibly, resulting in more accurate and comprehensive outputs.
The model achieved an average $F_1$ of 88\% (soft match score, defined in \textit{Methods}) across various benchmarks for the reaction component identification task, significantly outperforming the state-of-the-art method by an average of 5\%. Additionally, our tests highlighted the model's superior abilities in interpreting textual information that describes reaction conditions.

We further developed a web application that can easily be used and deployed. The web application is hosted at \url{https://huggingface.co/spaces/CYF200127/RxnIM}. The source code and data are available at \url{https://github.com/CYF2000127/RxnIM}.
Since RxnIM was trained on a promising large-scale data and offered as a ready-to-use open-source tool, we believe it will greatly reduce the workload and enable the construction of high-quality datasets for the research community and promote machine-learning-driven innovations in organic chemistry.

\section{Results}
\label{sec:exp}
\begin{figure*}[ht]
\centering
\includegraphics[width=1.0\textwidth]{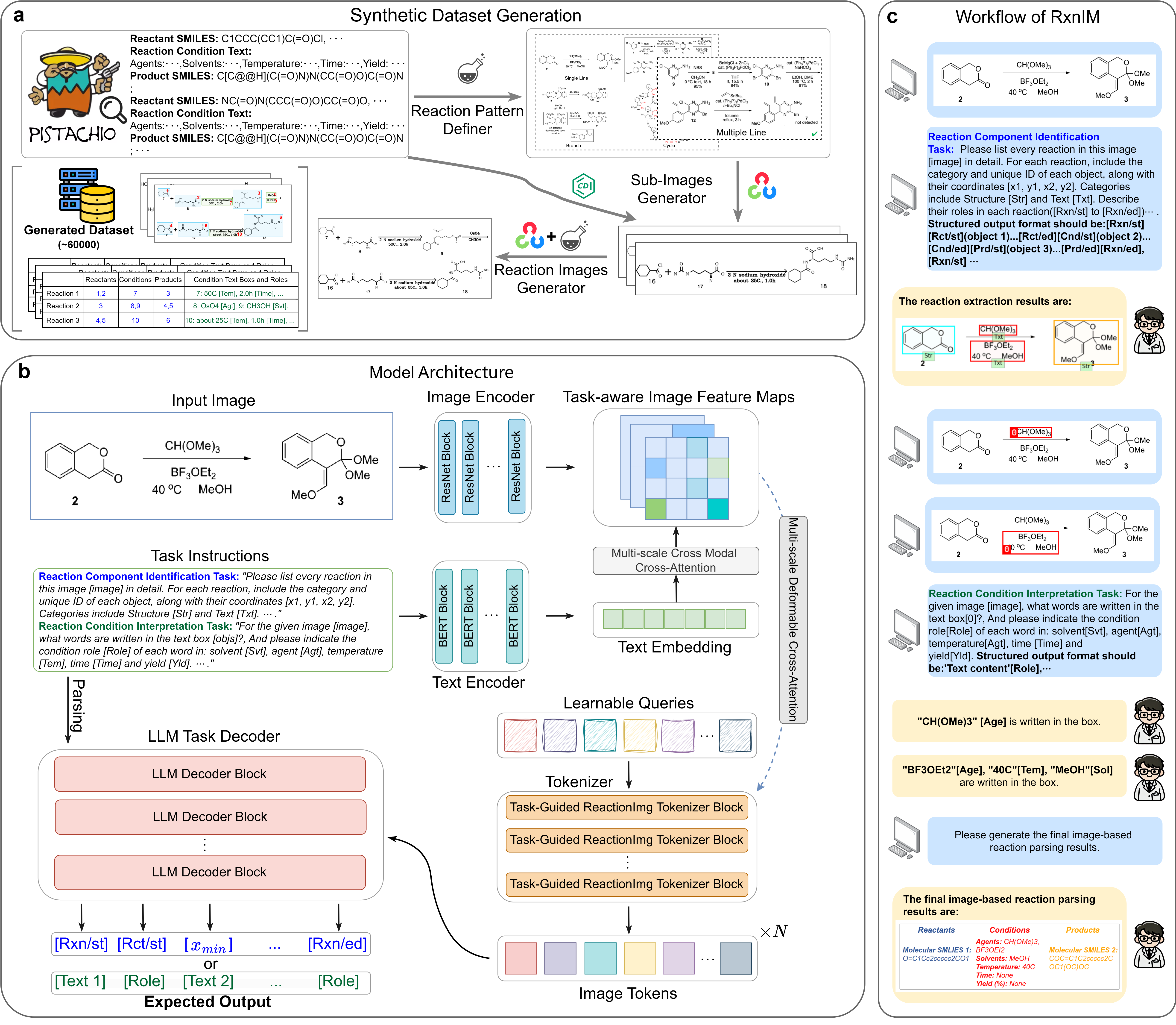}
\caption
{\textbf{Dataset generation and overview of the proposed RxnIM.} 
\textbf{a}, Synthetic dataset generation pipeline. We obtain textual reaction information in the Pistachio dataset, generate visual reaction components, and create sub-images based on predefined reaction patterns. These sub-images are combined to form the final synthetic reaction image. This process resulted in the creation of a large-scale chemical reaction image parsing dataset containing 60,000 diverse images.
\textbf{b}, Model architecture of our RxnIM. The model incorporates four key components: 
1) A unified task instruction for standardizing chemical reaction image parsing tasks,
2) A multimodal encoder that aligns image information with task instructions,
3) A ReactionImg tokenizer to convert image features into tokens, and
4) An open-ended LLM decoder that generates the final output.
\textbf{c}, Workflow for chemical reaction image parsing using RxnIM, where results from two tasks are combined and molecular structures are converted into machine-readable formats like SMILES or Molfile.}                        
\label{fig:rim}
\end{figure*}
% ------------------------------------------
In our workflow, the chemical reaction image parsing task is divided into two sub-tasks: reaction component identification and reaction condition interpretation. The reaction component identification task involves identifying all the reactions, segment their components and understanding their roles (such as reactant, condition, or product) in a reaction image. The reaction condition interpretation task is to extract the detailed condition in a reaction by recognizing the words in the text regions that describe reaction conditions and understanding their meanings (e.g., names of reagents or solvents, temperature, time, and yield). Further details of the task design can be found in the Methods section.

To minimize the effort for data labeling, we primarily used "synthetic data" to train our models. Specifically, we use structured reaction data from a large-scale chemical reaction database, Pistachio~\cite{15mayfield2018pistachio}, to construct images of chemical reactions following the general rules and styles in the chemical literature. For example, for a single step reaction, we first draw the images of reactant and product molecular structures using cheminformatics tools, and then place them on a canvas and draw an arrow that points from the reactant to the product. Reagent information is placed above the arrow, while solvent, temperature, reaction time and yield are placed below the arrow. In this way, we can construct a large number of labeled images of chemical reactions automatically. To account for the complexity of real chemical reaction images, we performed data augmentation in font size, line width, size of the molecular images, and reaction pattern (e.g., single-line, multiple-line, branch, and cycle). Full descriptions of the image generation and examples of generated images are available in the Methods section, and further details of the algorithm are available in Supplementary Method 1. 
Using this approach, we generated 60,200 synthetic images along with their corresponding ground truth data as our primary dataset. For each image, ground truth data includes the positions and roles of the reaction components, as well as the reaction condition texts. We divided the data into training, validation, and test sets using an 8:1:1 ratio.  To ensure the model performs well on a broader range of reaction images found in real literature, we also incorporated a small-scale real reaction image dataset manually labeled by Qian et al.~\cite{5qian2023rxnscribe}. We followed the original split of the real data set and used it in both the training and testing phases for the reaction component identification task. 
The details of each dataset can be found in Supplementary Note 1.

The training process was conducted in three stages, each utilizing different datasets and tasks. In the first stage, the model was trained on the synthetic dataset, focusing on the object detection task to accurately locate objects within the reaction images. In the second stage, the model was further trained on the synthetic dataset, incorporating both the reaction component identification and reaction condition interpretation tasks to enhance its ability to understand the roles and contents of the parsed objects. In the final stage, the model was fine-tuned using the real reaction image dataset specifically for the reaction component identification , allowing it to adapt to the more diverse and complex scenarios present in real-world chemical literature. The Implementation details can be found in the Methods section.

%%train 
%%corresponds to two generated ground truth data for the reaction component identification task and conventional object detection task, respectively.

%\subsection{Results of Different Reaction Data Extraction Tasks}
\subsection{Performance on the Reaction Component Identification Task}

For the reaction component identification task, we compared our RxnIM with current reaction component identification methods including rule-based OChemR~\cite{OChemR} and ReactionDataExtractor~\cite{3wilary2021reactiondataextractor}, and deep learning-based models RxnScribe~\cite{5qian2023rxnscribe}, using evaluation metrics including on precision, recall, and $F_1$ score. We adopted the same concepts of "hard match" and "soft match" as described in the RxnScribe paper, where a hard match only count instances where the prediction matches the ground truth exactly, while a soft match allows the labeling the role of a reagent as a reactant. \begin{table}[t]
  \centering
  \caption{\textbf{Overall comparison of model performance on the reaction component identification task on different test datasets.} We contrast the performance of RxnIM with other models on both synthetic and real datasets. We present detailed metrics for hard match and soft match criteria, including precision, recall, and $F_1$ scores. Scores are all in \%.}
  \label{tab:sota}
  \renewcommand\arraystretch{1.2}
  \setlength{\tabcolsep}{5pt}

  \begin{tabular}{@{}cccccccc@{}}
    \toprule
    \multirow{2}{*}{Dataset} & \multirow{2}{*}{Model} & \multicolumn{3}{c}{Hard Match} & \multicolumn{3}{c}{Soft Match} \\
    \cmidrule(lr){3-5} \cmidrule(lr){6-8} 
     & & Precision & Recall & $F_1$ & Precision & Recall & $F_1$ \\
    \midrule
    \multirow{4}{*}{Synthetic} & ReactionDataExtractor~\cite{3wilary2021reactiondataextractor} & 8.4 & 6.9 & 7.6 & 22.6 & 11.4 & 15.2 \\
     & OChemR~\cite{OChemR} & 8.1 & 7.1 & 7.8 & 15.9 & 12.8 & 14.2 \\
     & RxnScribe~\cite{5qian2023rxnscribe} & \underline{78.5} & \underline{75.6} & \underline{77.1} & \underline{87.6} & \underline{84.0} & \underline{85.8} \\
     & RxnIM & \textbf{86.4} & \textbf{85.9} & \textbf{86.2} & \textbf{91.6} & \textbf{90.8} & \textbf{91.2} \\
     \midrule
    \multirow{4}{*}{Real} & ReactionDataExtractor~\cite{3wilary2021reactiondataextractor} & 4.1 & 1.3 & 1.9 & 19.4 & 5.9 & 9.0 \\
     & OChemR~\cite{OChemR} & 4.4 & 2.8 & 3.4 & 12.4 & 7.9 & 9.6 \\
     & RxnScribe~\cite{5qian2023rxnscribe} & \underline{72.3} & \underline{66.2} & \underline{69.1} & \underline{83.8} & \underline{76.5} & \underline{80.0} \\
     & RxnIM & \textbf{74.7} & \textbf{69.7} & \textbf{72.1} & \textbf{86.9} & \textbf{82.8} & \textbf{84.8} \\

    \bottomrule
  \end{tabular}
 \vspace{-0.5cm}
  %\footnotetext{We contrast the performance of RxnIM with other models on both synthetic and real datasets. We present detailed metrics for hard match and soft match criteria, including precision, recall, and $F_1$ scores, highlighting RxnIM's superior performance, especially in challenging real-world data scenarios. Scores are all in \%.}
\end{table}

The results on the synthetic test dataset and the real test dataset are shown in Table~\ref{tab:sota}. RxnIM demonstrates better performance on various metrics when compared with other methods. Specifically, In the soft match criteria, RxnIM achieves a precision of 91.6\%, a recall of 90.8\%, and an $F_1$ score of 91.2\% on the synthetic test dataset, outperforming the second best method by 4.0\%, 6.8\% and 5.4\%, respectively. On the real data set where the images are more diverse and complex, RxnIM still reports a precision of 86.9\%, a recall of 82.8\%, and an $F_1$ score of 84.8\%, outperforming the second best method by 3.1\%, 6.3\% and 4.8\%, respectively. This indicates the advanced abilities of our model in extracting reactions in diverse reaction images, underscoring its robustness and adaptability to different levels of image complexity and variability. 

Under the hard match criteria, RxnIM reaches a precision of 86.4\%, a recall of 85.9\%, and a $F_1$ score of 86.2\% on the synthetic test dataset, outperforming the second best method by 7.9\%, 10.3\% and 9.1\%, respectively, and a precision of 74.7\%, a recall of 69.7\%, and an $F_1$ score of 72.1\% on the real test dataset, surpassing the second best method by 2.4\%, 3.5\% and 3.0\%, respectively. This further demonstrates RxnIM's robust and general capability to precisely and consistently extract the exact role of reaction components in diverse reaction images. 
\begin{figure*}[t]
\centering
\includegraphics[width=1\textwidth]{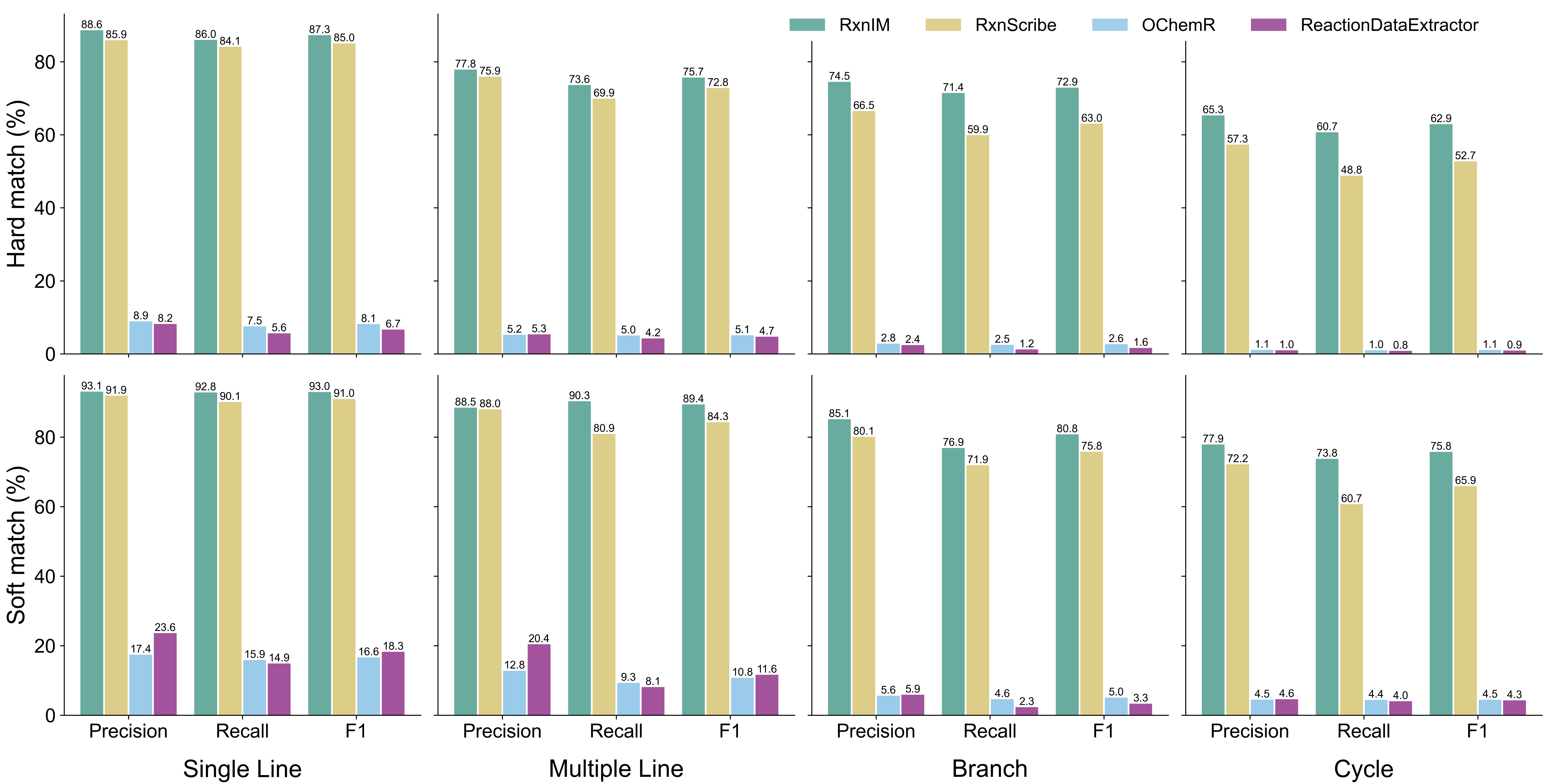}
\caption{\textbf{Comparison of model performance on the reaction component identification task on four different patterns of reaction images on the real test dataset.} We display precision, recall, and $F_1$ scores in hard match and soft match, of our model and current methods across four patterns of reaction images: Single Line, Multiple Line, Branch, and Cycle. The performance is evaluated to demonstrate the models' capabilities in accurately extracting reactions under varying image complexities and layouts.}                          
\label{fig:multi_results}
\end{figure*}
We further broke down the performance comparison into four different patterns of reaction images on the real test dataset - 1) Single line, where all reactions appear in the same line; 2) Multiple line, where there are multiple lines of reactions, 3) Branch, where branch is used when multiple reactions start from a common reactant, and 4) Cycle, where multiple reactions are displayed in a cycle, as shown in Fig.~\ref{fig:multi_results}. A detailed data distribution for four reaction image patterns is shown in Supplementary Note 2 and Supplementary Table 2.
For single-line reaction images, RxnIM achieves a hard match $F_1$ score of 86.0\%, while RxnScribe scores 84.1\%. In multiple-line images, RxnIM outperforms RxnScribe with a hard match $F_1$ score of 76.7\% versus 74.7\%. For branch images, RxnIM scores 71.4\% in hard match $F_1$, compared to RxnScribe’s 63.0\%. In cycle images, RxnIM achieves a hard match $F_1$ score of 60.7\%, while RxnScribe score are 52.7\%. RxnIM consistently outperforms RxnScribe across all reaction image patterns using both hard and soft match criteria. Meanwhile, the performance of rule-based methods remains at a relatively low level. The lower accuracy in multiple-line, branch, and cycle images is expected due to their diversity, complexity, and containing more reactions. However, the gap between RxnIM and RxnScribe widens in these categories, demonstrating our model's advanced image reasoning and localization abilities to handle more complicated reaction image patterns. We further compare and discuss the performance of these methods on four different patterns of reaction images on the synthetic test dataset in Supplementary Discussion 1 and Supplementary Table 5.

%%%%%%%%%%%%%%%%%%%% vis
\begin{figure*}[ht]
\centering
\includegraphics[width=1\textwidth]{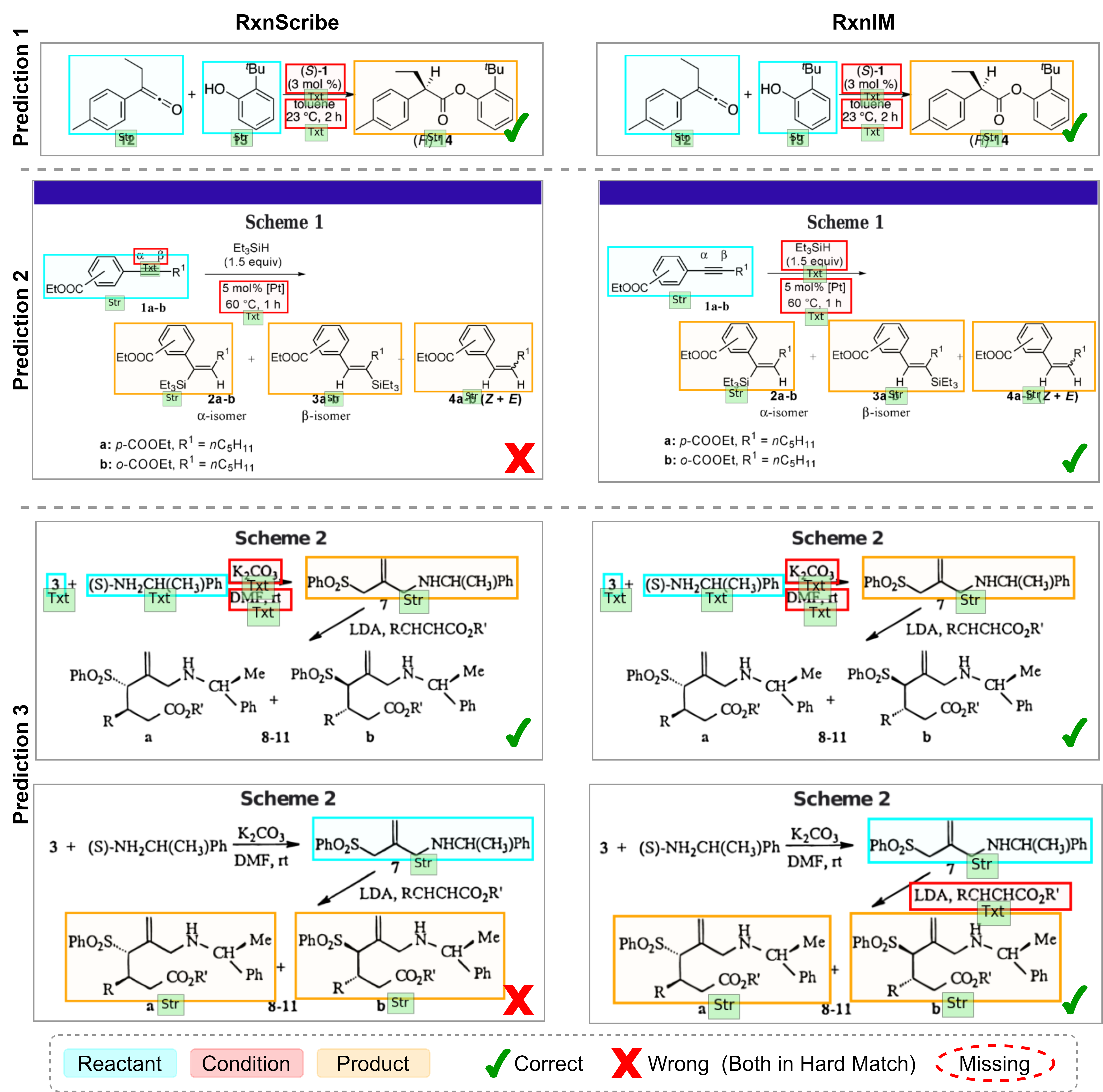}
\caption{\textbf{Visualization examples of the model's prediction on the reaction component identification task compared to the current best method RxnScribe.} We display the comparison between RxnScribe and RxnIM on the reaction component identification task across three different prediction examples. Each predicted reaction is visualized in a separate image, showing the predicted reaction components, including reactants, conditions, and products, with color-coded boxes representing different component types. Check marks and cross marks indicate correct and incorrect predictions, respectively, under the hard match criteria. The red dashed circle indicates that the reaction is not predicted. The DOI numbers of the relevant journal articles for these real reaction images can be found in Supplementary Note 3 and Supplementary Table 3.}                          
\label{fig:vs1}
\end{figure*}
In addition to quantitative measures, we show some examples of the reaction component identification task in Fig.~\ref{fig:vs1} to provide intuitive illustrations on the improvement achieved by RxnIM.
RxnScribe and RxnIM both make the correct prediction in prediction 1, which is a simple single-line image.\begin{figure*}[t]
\centering
\includegraphics[width=0.95\textwidth]{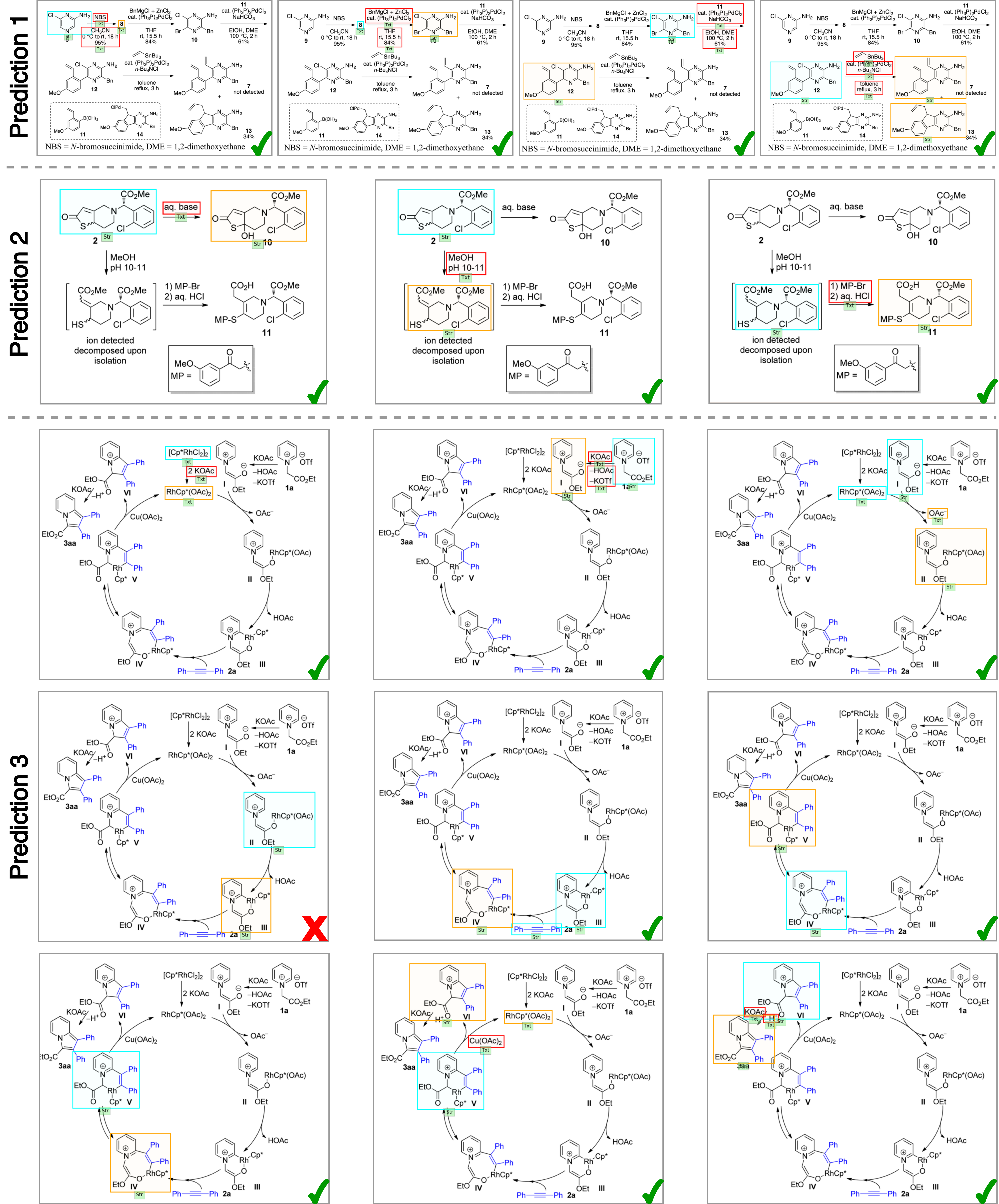}

\caption{\textbf{More visualization examples of the model's prediction on the reaction component identification task.} We showcase more complex examples of predicted reactions, each visualized in separate images. Prediction 1 is a multiple-line reaction image with four reactions, Prediction 2 is a branch reaction image with three reactions, and prediction 3 is a cycle reaction image with nine reactions.}                          
\label{fig:more_vis}
\end{figure*} 
In prediction 2, RxnScribe misinterprets several key objects in reactants and conditions, leading to inaccuracies in the reaction representation. This is due to this diverse single-line image containing a complex molecular structure and being line-wrapped when placing products. For this example, RxnIM accurately depicts the relationship between molecules in different lines and clearly labels the relevant reaction conditions. In prediction 3, which is an unusual multiple-line image with a nonparallel arrow surrounded by condition text, RxnScribe makes mistakes in labeling conditions in the second reaction step. RxnIM, however, successfully decodes all reaction steps, showcasing its robustness in handling complex reaction images.

%%%%%%%%%%%%%%%%%%%%%%%%

We further show some RxnIM's prediction on more complex reaction images in Fig.~\ref{fig:more_vis}.
Prediction 1 is a multiple-line image with four reactions. RxnIM correctly predicts all reactions, even when the products of the last reaction are placed vertically which is a rare style.
In Prediction 2, which is a branch image containing two different branches and three reactions in total, and RxnIM correctly identifies all reactions in each branch. 
Prediction 3 is catalytic cycle which represents one of the most complex styles in reaction images. This image contains nine diverse reactions, with a blend of many different types of arrows such as curved, branching, vertical, and bidirectional. RxnIM successfully makes eight correct predictions out of nine reactions. The only incorrect prediction is a reaction where the model misses a small-molecule byproduct, and the major reactants and products are well recognized. 

The performance of RxnIM across various metrics, datasets, and reaction image patterns underscores its effectiveness in handling the complexities of the reaction component identification task in the chemical literature, highlighting its exceptional image reasoning and localization abilities to extract every reaction step in complex and diverse reaction images.

\subsection{Performance on the Reaction Condition Interpretation Task}

RxnIM uniquely leverages the multimodal capabilities of LLMs to perform multiple tasks within a unified framework. By integrating visual and textual information, our model is able to not only identify the locations of reaction components, but also interprate the textual content in the components, i.e. perform the reaction condition interpretation task, differentiating it from existing models.

\begin{figure*}[t]
\centering
\includegraphics[width=1\textwidth]{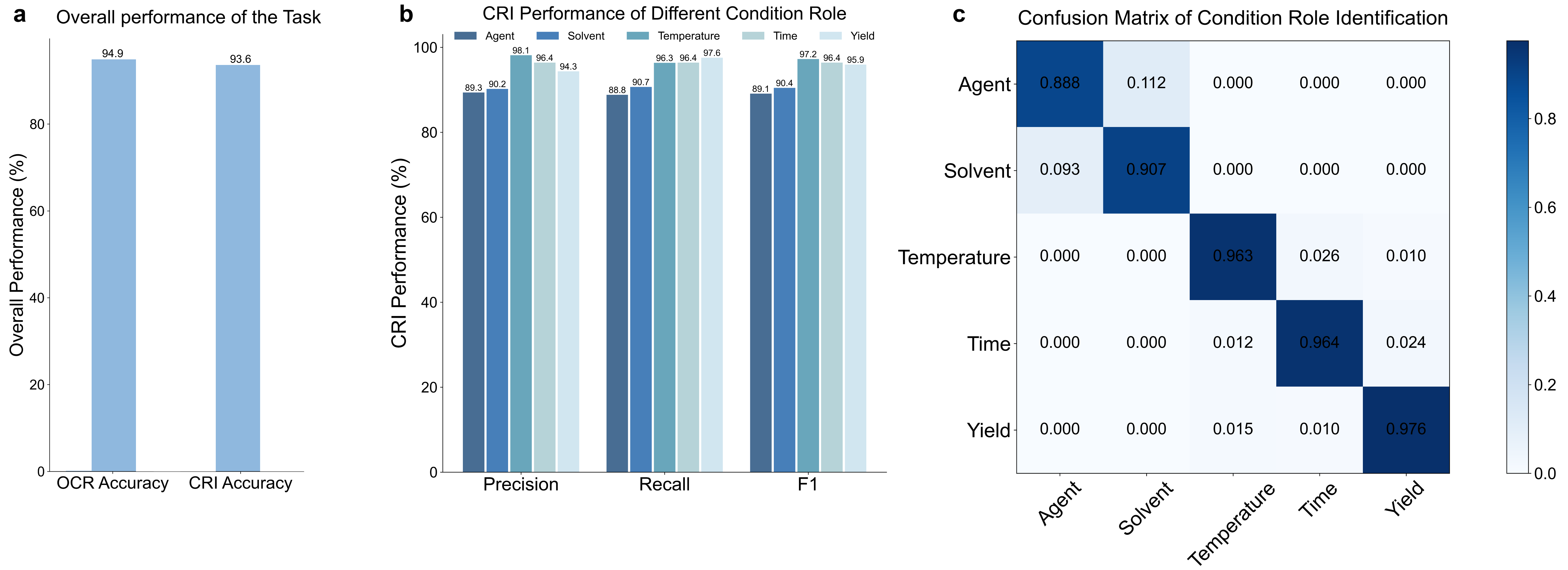}
\caption{
\textbf{Model performance on the reaction condition interpretation task.} 
\textbf{a}, overall performance on the reaction condition interpretation task in OCR and CRI (Condition Role Identification) accuracy.
\textbf{b}, the CRI performance in precision, recall, and $F_1$ scores on five different condition roles: agent, solvent, temperature, time, and yield.
\textbf{c}, the confusion matrix detailing the model's performance in correctly identifying these condition roles, highlighting areas of accurate and confused classifications.
}
\label{fig:OCR}
\end{figure*}

The overall performance of RxnIM on the reaction condition interpretation task is outlined in Fig.~\ref{fig:OCR}(a). We evaluated the performance from two perspectives - 1) whether the text of the reaction condition is correctly recognized, which we term "condition OCR", and 2) whether the roles of different elements in the reaction condition are correctly understood, which we term "condition cole identification (CRI)". For condition OCR, RxnIM achieved a high accuracy of 94.9\%, indicating the model's effectiveness in recognizing and converting text within chemical images into editable and searchable data. 
For the CRI task, the model reached an accuracy of 93.6\%. The CRI performance in Fig.~\ref{fig:OCR}(b) and the confusion matrix in Fig.~\ref{fig:OCR}(c) further details the model’s performance across various condition roles such as reagent, solvent, temperature, time, and yield. The model demonstrates strong performance in identifying reagents with a precision of 89.3\%, a recall of 88.8\%, and an $F_1$ score of 89.1\%. For solvent, the precision is 90.2\%, recall is 90.7\%, and $F_1$ score is 90.4\%. These results suggest that the model effectively distinguishes between chemical agents and solvents. It is worth noting from the confusion matrix that reagents and solvents are more often misidentified with each other, as compared to with other elements. This is likely due to some inherent ambiguity of these two roles - some chemicals can both serve as the solvent to dissolve the reactants, and the reagent to promote reactivity. For numerical parameters like temperature, time, and yield, the model exhibits high accuracy. The precision, recall and $F_1$ scores are 96.3\%, 96.4\%, and 97.6\%, respectively, indicating strong performance in identifying and classifying these crucial elements in reaction conditions. The confusion matrix further confirms the model's accuracy, showing minimal misclassification among these categories.

\begin{figure*}[t]
\centering
\includegraphics[width=1\textwidth]{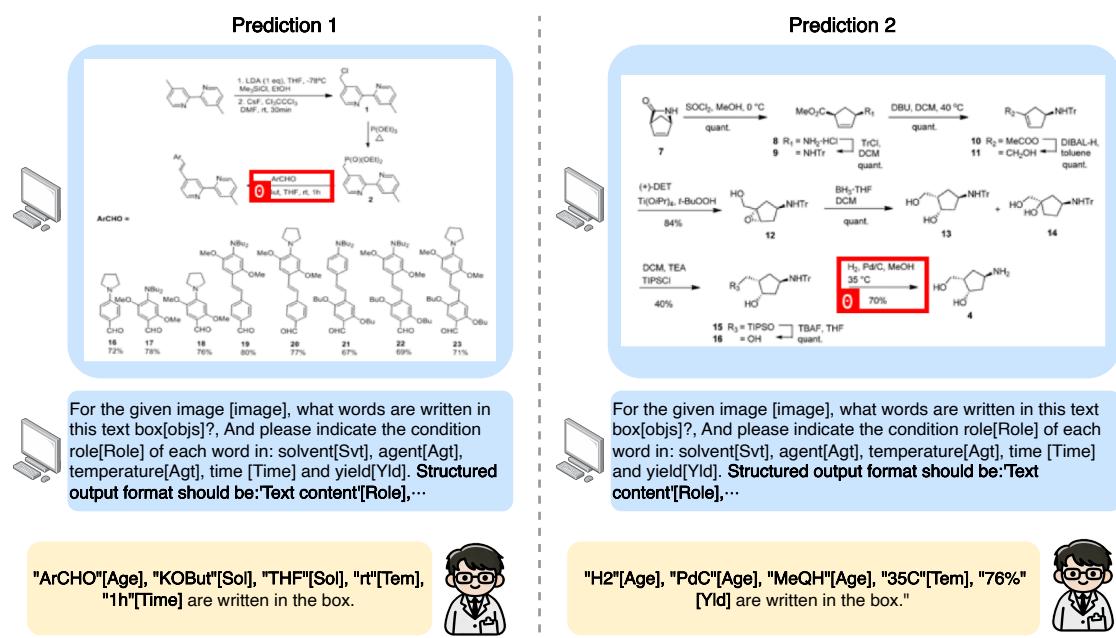}
\caption{\textbf{Visualization examples of the model’s prediction on the reaction condition interpretation task.} We show example outputs of the model for the reaction condition interpretation task across two different predictions. Each prediction involves extracting and identifying the text and corresponding roles within reaction condition boxes. The output format indicates the recognized text along with its assigned condition role, such as agent [Age], solvent [Sol], temperature [Tem], time [Time], and yield [Yld]. }                          
\label{fig:vs2}
\end{figure*}

We also visualize some examples of this task in Fig.~\ref{fig:vs2}.
RxnIM exhibits strong performance in accurately extracting and categorizing text within reaction conditions. In two separate predictions, it effectively identifies and assigns roles to chemical reagents, solvents, temperatures, and times. For instance, in the first prediction, it correctly labels 'ArCHO' as a reagent and 'THF' as a solvent, while in the second, it accurately recognizes 'H2' and 'PdC' as reagents, and '35C' as the temperature. These results underscore the model's precision and utility in parsing condition text information in reaction images, making the final reaction component identification results more detailed and comprehensive.

Overall, our comprehensive evaluation of RxnIM shows that it consistently outperforms the current methods. It demonstrates promising abilities in image reasoning, localization, and OCR, proving to be a reliable tool for chemical reaction image parsing and machine-readable reaction database construction. 

\subsection{RxnIM.web}
\begin{figure*}[ht]
\centering
\includegraphics[width=1\textwidth]{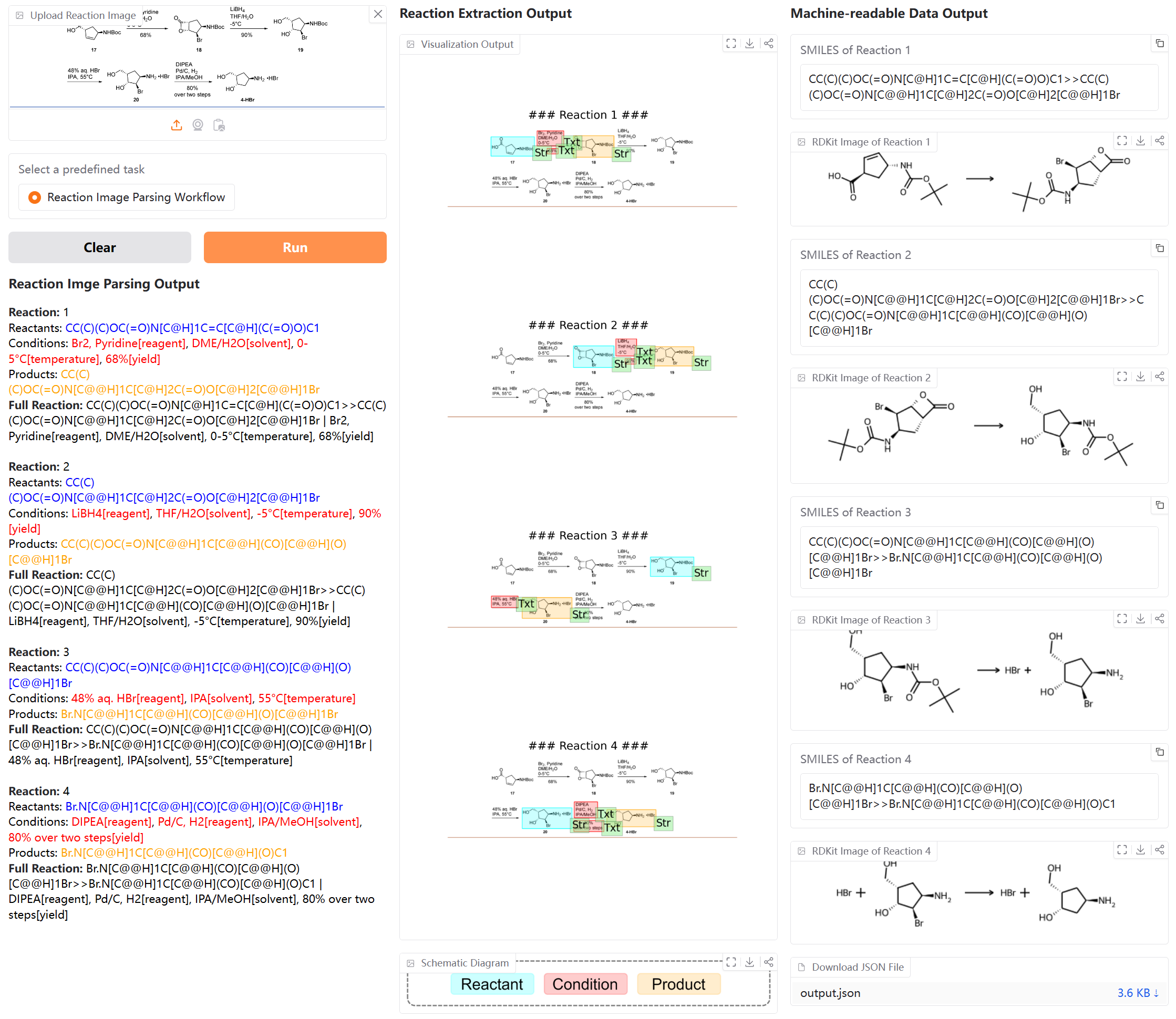}
\caption{\textbf{Example of a multiple-line reaction image (on the top left) that has been loaded into the RxnIM web application.} We visualize the output of reaction component identification task and integrate the detailed results based on reaction image pasring output into machine readable formats such as reaction SMILES string and json format.}                          
\label{fig:webapp}
\end{figure*}
RxnIM.web is a web application that combines the outputs of the previously described tasks in a straightforward, comprehensive workflow for reaction image parsing, as shown in Fig~\ref{fig:rim}(c). We also provide an example of RxnIM.web in action in Fig~\ref{fig:webapp}. In this case, the RxnIM.web begins with the user uploading a multiple-line reaction image. RxnIM.web will first run reaction component identification task to extract the regions of reactants, conditions and products for each reaction. Visualized outputs are presented in the \textit{Reaction Extraction Output
} section, each reaction is displayed individually in a image. Then RxnIM.web will run  reaction condition interpretation tasks for each condition text regions to extract the detailed conditions for each reaction. Molecular objects are then processed by a molecular recognition model to obtain machine-readable SMILES string. Finally, RxnIM.web will integrate this information in the \textit{Reaction Image Parsing Output} section.

Additionally, in the \textit{Machine-readable Data Output} section, we integrate the SMILES part of each reaction in the reaction image parsing output into reaction SMILES strings, making them accessible for common chemistry tools such as RDKit and ChemDraw. Each reaction structure is also visualized as A 2D molecular diagram by RDKit. Besides, the reaction image parsing output can also be downloaded in machine-readable json file format in this section.

\subsection{Effect of Model Components and Configurations}
\begin{figure*}[t]
\centering
\includegraphics[width=1\textwidth]{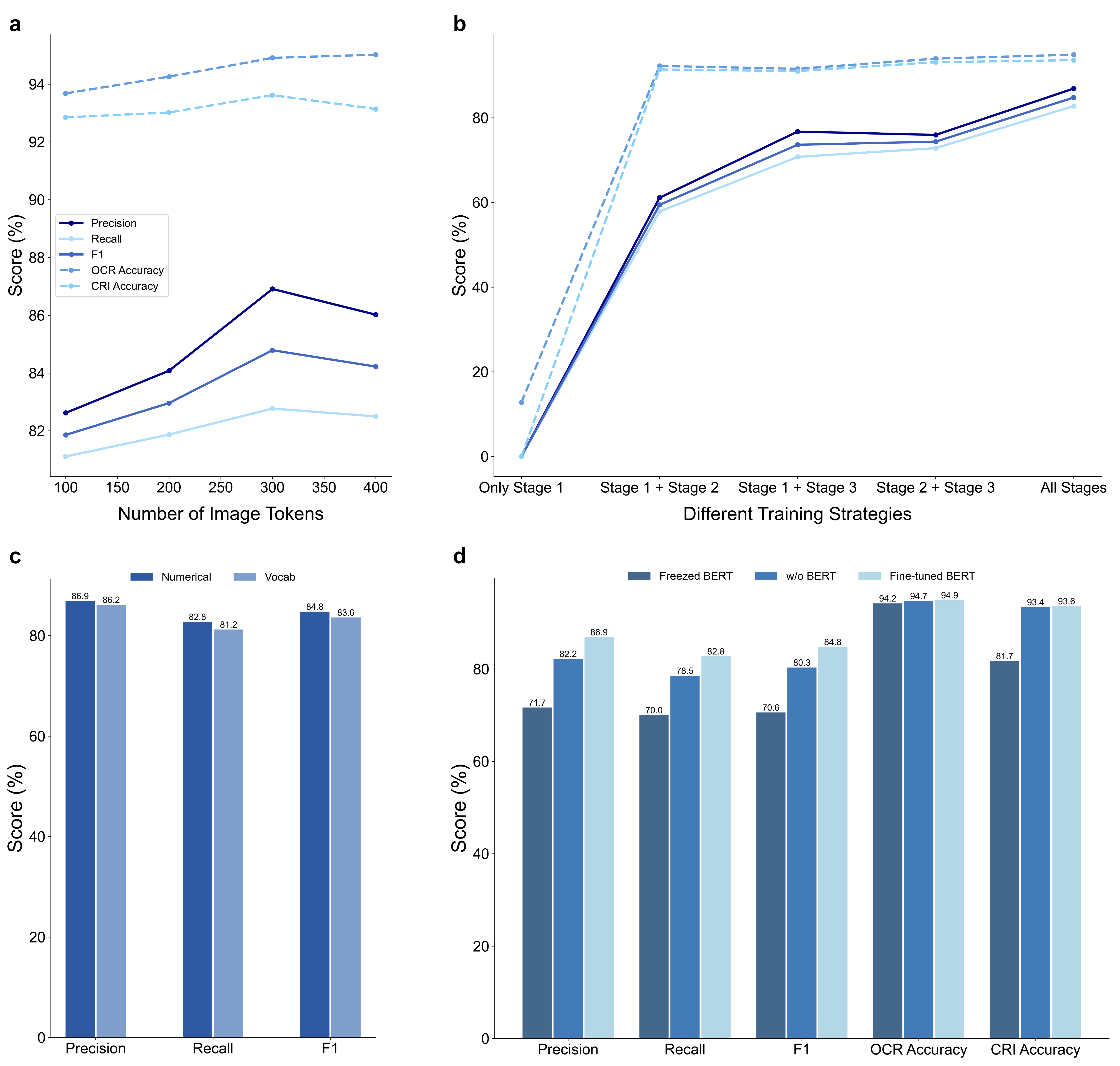}
\caption{\textbf{Model performance with different components and configurations}
\textbf{a}, effect of varying the number of image tokens in precision, recall, $F_1$ score, OCR accuracy, and CRI accuracy.
\textbf{b}, influence of training strategies using different training stages, showing performance improvements across multiple training stages.
\textbf{c}, compares the use of special location tokens versus numerical tokens in position representation.
\textbf{d}, effect of the text encoder, contrasting performance with a frozen BERT, without BERT, and with a fine-tuned BERT.
More discussions of model performance with different components and configurations are illustrated in Supplementary Discussion 2 and 3.}                          
\label{fig:abl}
\end{figure*}

In this section, we provide evaluations of the components and configurations that influence the performance of our model in terms of precision, recall, and $F_1$ under the soft match criteria for the reaction component identification task, as well as OCR accuracy and CRI accuracy. All evaluations for the reaction component identification task are conducted on the real test dataset. The results are shown in Fig~\ref{fig:abl}.

\subsubsection{Number of Image Tokens}
%\input{tables/tab_abl_tokens}
% We discuss the impact of using different numbers of Image Tokens on the performance of the model which is shown in Fig~\ref{fig:abl}(a). The results show that increasing the number of image tokens from 100 to 300 generally improves the performance across all metrics, peaking at 300 tokens with the highest scores in Soft Match $F_1$ (84.79\%), OCR Accuracy (94.91\%), and Role Identification Accuracy (93.62\%).  However, increasing the number to 400 does not continue this trend and sees a slight decrease in most scores, indicating that there might be an optimal number of tokens around 300 that balances performance without overfitting or underutilizing the model's capacity. Therefore, we choose 300 image tokens in our task-guided ReactionImg tokenizer.
In our exploration of the effect of varying the number of image tokens, depicted in Fig~\ref{fig:abl}(a), we observe a clear trend of increasing performance across all metrics as the number of tokens increases from 100 to 300. The performance peaks at 300 tokens, achieving a soft match $F_1$ score of 84.8\%, OCR accuracy of 94.9\%, and CRI accuracy of 93.6\%. This improvement suggests that a higher number of tokens allows the model to capture more detailed information from the images, enhancing its ability to differentiate and recognize complex structures and reaction components. However, as the number of tokens exceeds 300, we observe a slight decline in performance at 400 tokens. This decrease may be due to the reduced effectiveness of capturing long-range dependencies between tokens, as more tokens could represent smaller image regions, leading to diminished global context understanding. This suggests that 300 tokens provide an optimal balance, capturing sufficient detail while maintaining effective long-range relationships. Therefore, we select 300 image tokens as the ideal configuration to maximize performance without compromising the model's ability to understand the overall image structure.

\subsubsection{Influence of Training Strategies}
The analysis of training strategies using different training stages is shown in Fig~\ref{fig:abl}(b). Detailed training strategies are described in Methods section and 
Supplementary Notes 4.
Starting with the use of the single first training stage, the model cannot effectively perform the two tasks because this stage only pre-trains the vision part to perform the object detection task. As the second stage is incorporated, there is a noticeable improvement in the scores. Specifically, when the reaction component identification task and  reaction condition interpretation task are involved, the model's performance metrics see a marked increase, suggesting that the tasks we designed significantly enhance the model's ability to parse reaction data. 
The most substantial gains are observed when all three stages are integrated, indicating the cumulative benefits of a comprehensive training strategy. The results from these stages indicate that the sequential and targeted training strategy significantly enhances the model's performance. Each stage builds upon the previous one, progressively refining the model's capabilities in localization, recognition, and image reasoning sequentially. The combination of pre-training, comprehensive fine-tuning, and focused adaptation to real data ensures that the model not only excels in synthetic scenarios but also demonstrates strong generalization and robustness in practical applications. 

\subsubsection{Special Location Tokens vs. Numerical Tokens}
\label{sec:token}
We compared the different position representation techniques in Fig~\ref{fig:abl}(c). Detailed position representation is described in Methods section. The model using numerical representations outperforms that using vocabulary-based representations. This advantage likely arises from the numerical method's ability to provide more granular and precise positional information, which is critical in understanding spatial relationships in reaction images. Numerical tokens can encode exact coordinates and sizes, enabling the model to better differentiate between closely positioned elements and capture the detailed layout of reaction schemes. While numerical tokens might increase computational complexity during training, the trade-off is justified by the substantial improvement in model accuracy and reliability. This finding underscores the importance of precise positional representations in enhancing a model's image reasoning and localization capabilities.

\subsubsection{Influence of Text Encoder}
We investigated the role of the text encoder BERT in the multimodal encoder in Fig~\ref{fig:abl}(d). The result illustrates that integrating BERT into the model without freezing any parameters provides a significant improvement in all metrics compared to configurations without BERT or with BERT and frozen parameters. This suggests that BERT's powerful contextual understanding of the task instructions significantly contributes to the model's performance, particularly when it is allowed to adapt to the specific context of the reaction parsing tasks. Freezing parameters while using BERT results in a noticeable drop in performance, particularly in CRI accuracy and soft match $F_1$, underscoring the importance of dynamic parameter adjustment during training.

% \subsection{Qualitative Results}
% We assess the performance of our RxnIM through specific test cases. This qualitative analysis aims to further validate its effectiveness and comprehensive capabilities in extracting data from diverse reaction data extraction tasks.
% \subsubsection{Reaction Component Identification Task}

% We first conduct an ablation study to evaluate the impact of multi-task learning using task instructions. As shown in Fig~\ref{fig:abl}(b), the results demonstrate minimal differences between models trained with single task or two tasks together. Both configurations achieve similar performance metrics across all evaluated categories, the performance of Joint training even slightly better than that of separate training, suggesting that the model's architecture is robust enough to handle joint or separate training without significant impact on the outcomes, and the multiple tasks training enhances the performance of the model.

\section{Discussion}
% In this study, we introduced RxnIM, a Multimodal Large Language Model (MLLM) tailored for extracting data from reaction images.  This model leverages advanced MLLM methodologies to achieve superior performance in tasks such as reaction extraction, condition OCR, and role identification.  By integrating image reasoning, localization, and OCR functions, RxnIM not only achieves different functions on a single model but also performs far better than traditional models. Experiments on different test sets further demonstrate its robustness.  The effective training of RxnIM was supported by a novel algorithm that generated a large dataset of synthetic reaction images, supplemented with real annotated images.  This comprehensive approach enabled the model to adapt to the diverse styles found in the chemical literature, significantly enhancing its performance and establishing a new standard in the field of chemical reaction data extraction.  Future work will aim to expand the model’s applications to broader chemical imaging contexts.
RxnIM employs advanced MLLM methodologies to achieve better performance in reaction component identification and reaction condition interpretation tasks. By integrating different tasks into a unified framework, RxnIM is able to perform multiple tasks within a single architecture, streamlining the process of reaction component identification and condition interpretation . The robustness and versatility of RxnIM were validated through extensive experiments on diverse test sets. In the reaction component identification task, we observe that for simple and well-structured reaction images, both RxnIM and the current best method achieve an $F_1$ score nearing 90\%. As the complexity of the reaction images increases, the performance of the existing method declines more sharply, while RxnIM is able to maintain strong results (see Fig~\ref{fig:multi_results} and Fig~\ref{fig:vs1}). This robustness is attributed to MLLM's advanced image reasoning and localization capabilities, which allow it to effectively parse intricate and diverse reaction images. Furthermore, RxnIM is the first model to successfully implement the recognition and classification of reaction condition texts, enhancing its ability to handle both structural and contextual elements of chemical reactions, making the final extraction results more comprehensive.

Another key contributor to the model's strong performance was the use of the large set of synthetic reaction images. This synthetic database allows RxnIM to generalize beyond the limited examples of manually curated real data.  Synthetic images are crucial for training because they offer a wide range of reaction scenarios. This approach also reduces the need for extensive manual annotation, a major bottleneck in building high-quality reaction datasets. By combining the synthetic data with a small but diverse set of manually labeled real reaction images, we ensured that RxnIM could maintain high performance even when applied to more complex real-world reaction images. 
The three-stage training process also played a critical role in optimizing model performance.  Initially focusing on conventional object detection to locate reaction components, then expanding to understanding their roles and parsing the reaction condition text information, and finally fine-tuning on real reaction images, the model was able to progressively build more complex capabilities. Experiments demonstrate that our structured training process allowed RxnIM to learn effectively from integrated synthetic and real data, resulting in a model that is not only accurate but also robust across a wide variety of reaction image formats and complexities (see Fig~\ref{fig:abl}(b)).

One limitation of RxnIM is its inability to directly output the molecular structures in the form of SMILES. The accurate generation of SMILES from visual molecular representations remains a significant challenge for MLLMs, as it requires precise chemical structure interpretation and conversion. This limitation underscores an area for future improvement, and the potential for enabling direct SMILES output in the future would significantly enhance the model's utility for synthetic chemistry applications.

Overall, our work provides a powerful tool for chemical reaction image parsing while also serving as a valuable data resource for the wider research community. Additionally, it opens up new avenues for applying multimodal large language models to broader image-based cheminformatics. Future work will aim to expand the model's ability to understand detailed chemical structures, such as directly parsing SMILES data from reaction images and understanding reaction mechanisms in it, further enhancing the model's utility and scope.

% ------------------------------------------
\section{Methods}
\label{sec:method}
\subsection{Synthetic Dataset Generation}
\begin{figure*}[t]
\centering
\includegraphics[width=1\textwidth]{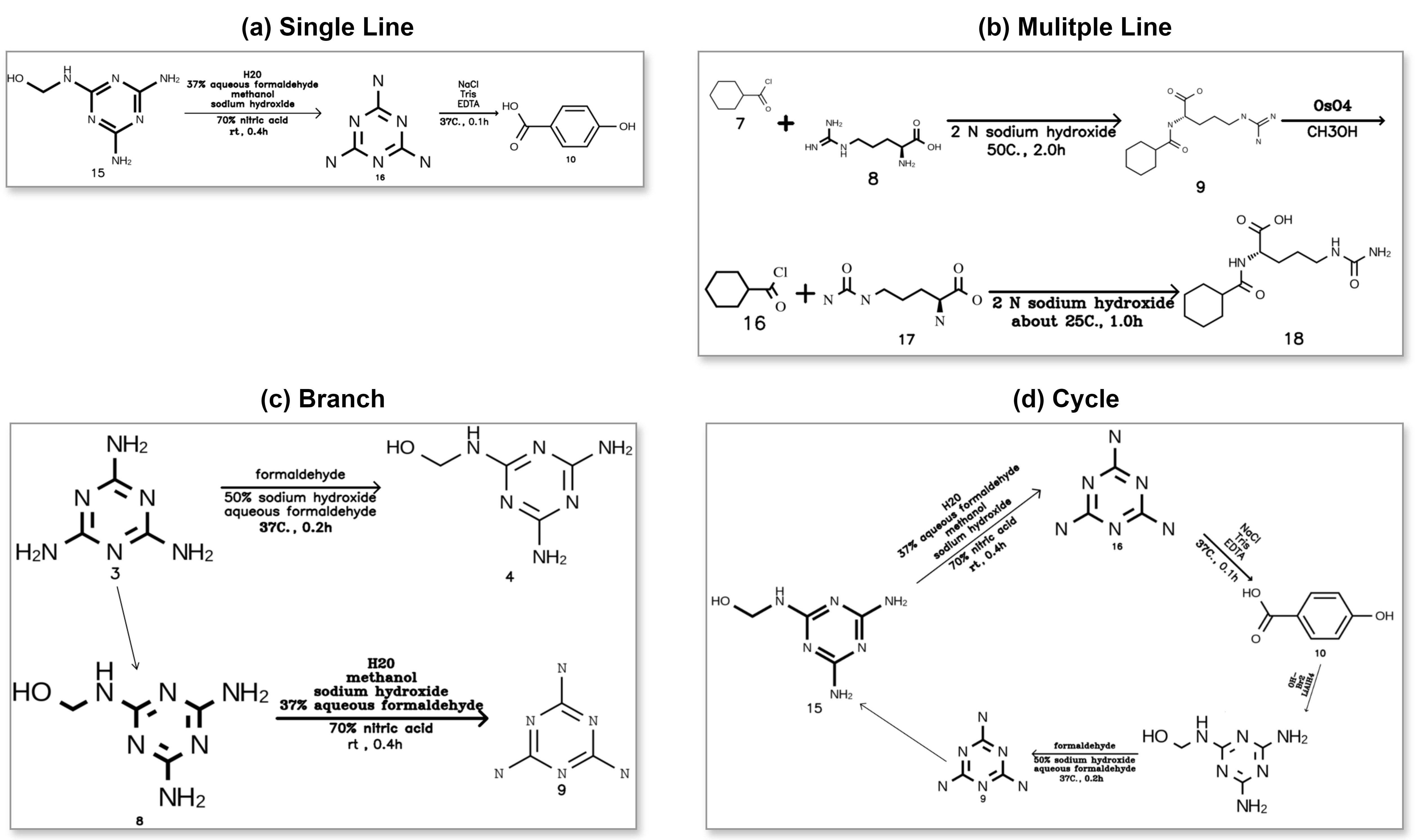}
\caption{\textbf{Examples of reaction images in four different patterns generated by our synthetic image generation algorithm.} We showcase the diverse types of reaction images produced by our algorithm, illustrating the range of patterns, including (a) Single Line, (b) Multiple Line, (c) Branch, and (d) Cycle. Each pattern represents a different layout and complexity of real chemical reactions, demonstrating the capability of our algorithm to create a wide variety of reaction images for training and evaluation purposes.}                       
\label{fig:img_sys}
\end{figure*}

\label{sec:data}
To train the reliable RxnIM, we developed an algorithm to artificially create synthetic images similar to real-world reaction images automatically. The algorithm used structured reaction data to construct reaction images following general conventions in chemistry. We used the Pistachio dataset, which includes structured information of reactant and product SMILES, along with reaction conditions such as agents, solvents, temperature, time, and yield. Then we define the reaction patterns and layout rules in single-line, multiple-line, branch, and cycle images. A sub-image generator with OpenCV and Indigo toolkit then creates several sub-images with reaction components like molecular structures, reaction arrows, and condition texts, adhering to common layout rules in chemistry. Since we have complete control of the image  generation process, we can automatically record the ground truth for the prediction tasks (e.g. locations of the objects and their roles). These sub-images and ground truth data are synthesized into labeled reaction images  by the reaction images generator, which arranges them according to the predefined patterns and layout rules. We show some examples of synthetic images in four patterns generated by our algorithm in Fig.~\ref{fig:img_sys}. The detailed definition of reaction patterns, layout rules, and generation for different reaction patterns are presented in Supplementary Method 1. The algorithm generateed 60,200 synthetic images and corresponding ground truth, which effectively solved the problem of the lack of large-scale diversified chemical reaction image datasets and provided strong support for training the model.

\subsection{Language-based Task Instruction for Reaction Image Parsing}
\label{sec:Language-based Task Instruction}
We introduce the language-based task instructions for the reaction component identification task, and the  reaction condition interpretation task. This design can also be further used to customize more reaction component identification task instructions. We describe the tasks by providing a task description with placeholders and special reaction role tokens and specifying the desired output format via task instructions.

\subsubsection{Reaction Component Identification Task}
\label{sec:Reaction Component Identification Task}
We construct the reaction component identification task as an extension of object detection. Previous works~\cite{5qian2023rxnscribe, chen2021pix2seq,wang2024visionllm} represent each object as five tokens i.e., $\text{Object} = [x_{min}, y_{min}, x_{max}, y_{max}, class]$, where first four tokens describe its bounding box in the image, $(x_{min}, y_{min})$ and $(x_{max}, y_{max})$ are the coordinates of the top-left and bottom-right points, respectively. They also quantize the continuous image coordinates into extra discrete tokens by binning $(e.g., <bin_0>,\ldots,<bin_{999}>)$.
We made a few modifications to represent the object clearer and more accurately. First, We represent coordinates using numbers directly($e.g.$, use three single-digit original number tokens to represent a coordinate from 0 to 999), Sec~\ref{sec:token} further discusses the different performance between these two representations. Second, we allocate a unique ID token following the class token for each object, enabling the model to represent each object more accurately. Then the class token represents the object type i.e. $class = [\text{Str}]\ \text{or}\ [\text{Txt}]$. We define two classes of objects: molecule structure ($[\text{Str}]$) and text description ($[\text{Txt}]$). Usually, reactants and products are represented as molecular structures, and conditions are described in text in the image. Finally, each object in the reaction image is represented as $\text{Object}_i = [x_{min}, y_{min}, x_{max}, y_{max}, class, \text{ID}_i]$. This format is also used as the output in the conventional object detection task during the first stage of our training. We then extend this format into the reaction component identification output by following a specialized reaction sequence:
\begin{flalign}
&\ \text{Reaction}   =\textbf{[Rxn/st]}\quad \text{Reactant}\quad \text{Condition}\quad \text{Product}\quad \textbf{[Rxn/ed]},& \nonumber\\
&\  \text{Reactant}  =\textbf{[Rct/st]}\quad \text{Object}_i * n\quad \textbf{[Rct/ed]},\ (n > 0)& \nonumber\\
&\  \text{Condition} =\textbf{[Cnd/st]}\quad \text{Object}_i * m\quad \textbf{[Cnd/ed]},\ (m \geq 0)& \nonumber\\
&\  \text{Product}   =\textbf{[Prd/st]}\quad \text{Object}_i * n\quad \textbf{[Prd/ed]},\ (n > 0)& 
\nonumber
\end{flalign}

Where each reaction is a sequence starting with a $\text{[Rxn/st]}$ token, which consists of three reaction roles: Reactant, Condition, and Product, and ends with a $\text{[Rxn/ed]}$ token. Each reaction role is a subsequence of objects that begins with a start token ($\text{[Rct/st]}$,$\text{[Cnd/st]}$ or $\text{[Prd/st]}$), and ends with an end token ($\text{[Rct/ed]}$,$\text{[Cnd/ed]}$ or $\text{[Prd/ed]}$). The conditions $n > 0$ and $m \geq 0$ indicate that the Reactant and Product must contain at least one object. However, the Condition can be empty. 
Through the language-based task instruction, we ask the model to search all the reactions existing in the reaction image from the top left corner of the image according to this special reaction sequence, and finally output a set of reaction sequences.
An example of language task instruction for the reaction component identification is as follows: 

\textit{Please list every reaction in this image[image] in detail. For each reaction, include the category and unique ID of each object, along with their coordinates [x1, y1, x2, y2]. Categories include Structure ([Str]) and Text ([Txt]). Describe their roles in each reaction ([Rxn/st] to [Rxn/ed]), including Reactants ([Rct/st] to [Rct/ed]), Conditions ([Cnd/st] to [Cnd/ed]), and Products ([Prd/st] to [Prd/ed]). Note that Reactants and Products must include at least one object, while Conditions can be specified without any objects. Structured output format should be: [Rxn/st][Rct/st](object 1)$\cdots$[Rct/ed][Cnd/st](object 2)$\cdots$[Cnd/ed][Prd/st](object 3)$\cdots$[Prd/ed][Rxn/ed],[Rxn/st]$\cdots$. Only the Conditions section can be empty (i.e., [Cnd/st][Cnd/ed] without anything between).}

Where [image] is the Placeholder for the image token. The image tokens will replace it during training.

\subsubsection{ Reaction Condition Interpretation task}
\label{sec:Condition OCR Role Identification Task}
Reaction condition interpretation task is designed as an extension of scene text-centric visual question answering. For this task, our model only focuses on the regions that are recognized as the text description ($[\text{Txt}]$) of the condition in the image by the previous reaction component identification task. We define the reaction condition interpretation output format as follows:
\begin{equation}
\text{Condition Role} = [ \text{"Text"}\ [Role] ]* n,\ (n > 0)  \nonumber
\end{equation}
Where $\text{"Text"}$ represented the text recognition results of a single word and followed by a special condition role token to indicate its role in condition, i.e. $Role = [\text{Agt}], [\text{Svt}], [\text{Tem}], [\text{Time}],\text{or}\ [\text{Yld}]$. We define five classes of conditional roles that frequently occur in reactions: Agent($[\text{Agt}]$), Solvent($[\text{Svt}]$), Temperature($[\text{Tem}]$), Time($[\text{Time}]$), and Yield($[\text{Yld}]$).
We ask the model to identify all discrete words from a condition text region in the image and assign a condition role token to each word by task instruction.
We describe the language task instruction for the reaction condition interpretation as follows: 

\textit{For the given image[image], what words are written in this text box[objs]. And please indicate the condition role[Role] of each word in: solvent[Svt], agent[Agt], temperature[Agt], time [Time] and yield[Yld]. Structured output format should be:'Text content'[Role],$\cdots$.}

Where [objs] is the Placeholder for the corresponding text region. It will be replaced by the bounding box of the text region during training.

\subsection{Model Architecture}
\subsubsection{Multimodal Encoder}
\label{sec:encoder}
In the core model architecture, which is shown in Fig.~\ref{fig:rim}(b), we first propose a multimodal encoder that contains an image encoder and a text encoder to encode the given input image and the language-based task instructions into task-aware image features. 

Specifically, given a reaction image with height $H$ and width $W$, we feed it into an efficient image encoder ResNet~\cite{he2016deep} that consists of four ResNet modules, where each module has a scale dimension of 4. ResNet~\cite{he2016deep} captures multi-scale image feature maps $F_c^v$ with a spatial resolution of $H/4 \times W/4$, $H/8 \times W/8$, $H/16 \times W/16$, and $H/32 \times W/32$, respectively. 
Meanwhile, we utilize a text encoder BERT~\cite{devlin2018bert} to extract the text features $F_c^l$ from the language-based task instructions.
The text features are subsequently integrated across various scales of visual features via cross-attention mechanisms~\cite{vaswani2017attention}, producing multi-scale task-aware feature maps. This multimodal encoder facilitates the alignment of features between different modalities.

\subsubsection{Task-Guided ReactionImg Tokenizer}
\label{sec:Tokenizer}
Different from previous works~\cite{chen2023shikra,zhang2024gpt4roi,peng2023kosmos} that directly employ fixed-size image features or image patches as the image tokens that are fed into the LLM decoder, our approach incorporates a task-guided ReactionImg tokenizer to flexibly tokenize any number of image tokens that are seamlessly aligned with language-based task instructions.

Our task-guided ReactionImg tokenizer is built based on a transformer-based encoder-decoder network deformable DETR~\cite{zhu2020deformable}, to capture the high-level information of images. In the transformer encoder, a multi-scale deformable cross-attention module~\cite{zhu2020deformable} is used to exchange information among multi-scale task-aware feature maps. The encoder outputs are multi-scale task-aware feature maps with the same resolutions as the input. Both the key and query elements are pixels from the task-aware multi-scale feature maps. During the decoding process, we input these new feature maps as the key elements and randomly initialize $N$ learnable queries $Q = {\{q_i\}}_{i=1}^N$ to extract $N$ image tokens $T = {\{e_i,p_i\}}_{i=1}^N$. Each image token is represented by an embedding vector $e_i$ and a position vector $p_i$. The embedding vector $e_i$ includes the semantic information of the token, while the position vector $p_i$ represents its spatial position in the image.
This approach allows us to represent not only the objects within the reaction images but also to capture the spatial relationships between them. 
Furthermore, the visual representations extracted by this approach are highly informative for language-based task instructions. This means that our model is not only able to understand the content of an image but is also able to generate more targeted visual representations based on different task instructions.

\subsubsection{LLM Task Decoder}
\label{sec:Decoder}
Our task decoder is built based on a widely used LLM, Llama~\cite{touvron2023llama}, and is designed to handle various chemical reaction image parsing tasks with task instructions. However, Llama lacks the specialized tokens needed for these chemical reaction image parsing tasks. % Additionally, as a causal model, it is inefficient for chemical reaction image parsing tasks related to visual perception and positioning.
Therefore, we expand the vocabulary of Llama with special tokens designed for reaction parsing tasks as mentioned in Sec.~\ref{sec:Language-based Task Instruction}, including object tokens([Str], [Txt]), reaction token(([Rxn/st], [Rxn/ed]), reaction role tokens([Rct/st], [Rct/ed], [Cnd/st], [Cnd/ed], [Prd/st], [Prd/ed]), and condition role tokens($[\text{Agt}], [\text{Svt}], [\text{Tem}], [\text{Time}],[\text{Yld}]$). 
Furthermore, following~\cite{wang2024visionllm}, we apply the output-format-as-query decoding strategy to address the inefficiency of the causal LLM model. The LLM model first parses the task instructions into structured output format($e.g.$, "[Rxn/st] [Rct/st] $[x_{min}, y_{min}, x_{max}, y_{max}, class, \text{ID}_i]$ ... [Rct/ed] ... [Rxn/ed]" for reaction component identification task, "$\text{'Text'}[Role]$" for  reaction condition interpretation task). The tokens in the structured output format are then provided as queries to the decoder to generate the desired output based on the query. This approach enables the model to avoid inefficient token-by-token decoding when parsing chemical reaction images, particularly in the task involving visual perception and positioning. It strictly constrains the output to the specified structure, such as a reaction schema or condition role list, while maintaining a unified framework for these tasks.
Through this method, we convert these tasks into the token classification format, so that the cross-entropy loss can be applied to train and fine-tune the model.

% During the inference, the model first outputs the result of the reaction component identification task for the entire image. The results of the conditional OCR and role identification task are then output for . The final reaction data extraction results combine the outputs of two tasks.
% A molecular structure recognition model~\cite{qian2023molscribe} is then applied to convert the molecular object in bounding boxes into structured data such as SMILES~\cite{weininger1988smiles} or Molfile~\cite{dalby1992description}.

\subsection{Workflow of Chemical Reaction Image Parsing Using RxnIM}
After training the RxnIM model, we designed a workflow to obtain the final reaction image parsing results as shown in Fig.~\ref{fig:rim}(c). The model first outputs the result of the reaction component identification task for the entire image. Blue, red, and orange boxes represent reactants, conditions, and products in reaction roles, respectively, the categories of the objects are indicated in green boxes. Following this, it focuses on the regions identified as text description in condition, applying OCR and role identification to extract the information of condition roles. 

The results from both tasks are integrated, providing comprehensive data that combines structural and descriptive details of the reaction. Additionally, a molecular structure recognition model post-processes these results, converting visual molecular structures into machine-readable data formats like SMILES or Molfile. This conversion is crucial for the practical use of the data in synthetic chemistry applications.
The final output is structured data of the chemical reaction as shown in the bottom of Fig~\ref{fig:rim}(c), ready for documentation and computational analysis. 
%This workflow highlights our model’s capability to transform complex chemical reaction images into machine-readable data.

%\input{figures/algri}

\subsection{Evaluation Metrics}
\subsubsection{Reaction Component Identification Task}
Evaluating the results for the reaction component identification task is complex due to the nature of the predictions and the ground truth, which are sets of reaction structures. In many cases, the predictions may not align perfectly with the ground truth. For instance, the bounding boxes for entities might be slightly misaligned, or the predicted order of reactions might differ. Despite these discrepancies, many such instances should still be deemed correct.
Following~\cite{5qian2023rxnscribe}, We use two groups of evaluation metrics, hard match and soft match, to evaluate the model.

Specifically, We start by comparing a single predicted reaction, denoted as $\hat{R}$, with a single ground truth reaction, $R$. This involves establishing a mapping between the lists of objects in $\hat{R}$ and $R$. For each object in $\hat{R}$, we identify the corresponding entity in $\hat{R}$ that has the highest bounding box overlap. This overlap is quantified by the intersection over union (IoU) score. If the maximum IoU exceeds a threshold of 0.5, the predicted and ground truth bounding boxes are considered to have successfully matched.

Furthermore, a prediction $\hat{R}$ is considered a match with the ground truth $R$ in the hard match evaluation only if all reactants, conditions, and products between $\hat{R}$ and $R$ can be aligned perfectly. In contrast, the soft match evaluation only focuses on molecule objects and does not differentiate between reactants and agents, which are part of the conditions.
The rationale for using soft match evaluation is twofold. First, it considers only molecular objects and not textual objects, addressing the common question of how consecutive lines of text are annotated, e.g., viewed as a single object or multiple objects. Second, it avoids distinguishing between reactants and conditions. This is because, in certain cases, a molecule may appear above or below the reaction arrow, which is mainly used for placing reaction conditions, but functionally acting as a reactant due to its significant contribution of heavy atoms. This ambiguity about the separation between reactants and reagents is tolerable in chemistry.

In each reaction image, the ground truth is represented as $G_r=\{R_1, R_2,\ldots, R_n\}$, and the prediction is represented as $P_r=\{\hat{R}_1,\hat{R}_2,\ldots, \hat{R}_m\}$. Then we compute the precision, recall, and $F_1$ scores for hard match and soft match, respectively. Since we do not have a one-to-one correspondence between the predicted reactions and the ground truth reactions, we list all pairs and compare each $\hat{R}_i$ to each $R_j$. The metric is defined as follows:

\begin{equation}
\text{Precision} = \frac{1}{m} \sum_{j=1}^{m} | (\exists i \in \{1, \ldots , n\}, \hat{R}_j \text{ matches } R_i)
\end{equation}

\begin{equation}
\text{Recall} = \frac{1}{n} \sum_{i=1}^{n} | (\exists j \in \{1 , \ldots , m\}, R_i \text{ matches } \hat{R}_j)
\end{equation}

\begin{equation}
\text{$F_1$} = \frac{2 \cdot \text{Precision} \cdot \text{Recall}}{\text{Precision} + \text{Recall}}
\end{equation}
\subsubsection{ Reaction Condition Interpretation task}
For the  reaction condition interpretation task, we use OCR accuracy to determine the proportion of correctly identified characters across all characters. For the role identification, only words with an OCR accuracy above a threshold of 0.8 are included in the calculations.  The condition role identification (CRI) accuracy is then used to determine the proportion of words correctly classified by the model. Additionally, the conventional precision, recall and $F_1$ are utilized to analyze the performance of the model in each specific condition role. The OCR accuracy and CRI accuracy is defined as follows: 

\begin{equation}
\text{OCR accuracy} = \frac{\text{Number of Correct Characters}}{\text{Total Number of Characters}}
\end{equation}

\begin{equation}
\text{CRI accuracy} = \frac{\text{Number of Correct Predictions}}{\text{Total Number of Predictions}}
\end{equation}

\subsection{Implementation Details}
In our RxnIM, we employ ResNet-50~\cite{he2016deep} as the image encoder and BERT-Base~\cite{devlin2018bert} as the text encoder within our multimodal encoder framework. For the task-guided ReactionImg tokenizer, we adopt deformable DETR (D-DETR)~\cite{zhu2020deformable} to capture high-level information and extract image tokens. We set the number of queries and image tokens M to 300. And the number of encoder/decoder layers is set to 6 for D-DETR. Subsequently, we integrate Llama-2-7B~\cite{touvron2023llama2}, an advanced version of the Llama~\cite{touvron2023llama} model, as our LLM-based task decoder to handle different tasks effectively.

The training of the model is structured in three stages.  
In the first stage, We start by loading pre-trained weights for D-DETR, BERT, and Llama-2-7B. During this phase, we train the multimodal encoder and the ReactionImg tokenizer, while the LLM's parameters remain frozen. The focus is on conventional object detection task to enable the model first to locate objects in reaction images accurately.
In the second stage, all model parameters are fine-tuned. The training, which lasts for 30 epochs, involves the reaction component identification task and the  reaction condition interpretation task using the synthetic dataset split in a 7:3 ratio. The initial learning rate is set at $2 \times 10^{-4}$.
In the final stage, we only fine-tune the LLM while freezing all other parameters. The training utilizes the real-image dataset for the reaction component identification task, adjusting the data split to 8:2, focusing on parsing complex reaction image patterns over 50 epochs with an initial learning rate of  $2 \times 10^{-5}$.

We employ AdamW as the optimizer with a cosine annealing learning rate schedule throughout all training stages. The model is trained using 8 $\times$ NVIDIA H800 GPUs, ensuring robust computational support for our extensive training regimen. The complete training settings can be found in Supplementary Note 4.

\subsection{RxnIM.web}
The RxnIM.web application has been developed using Gradio, a Python-based web framework, and is hosted on Hugging Face Spaces.  This lightweight and flexible setup enables easy deployment through an intuitive web interface.

Upon launching, the web app initializes a server to handle user requests and preloads our RxnIM model for reaction image parsing and a molecular recognition model for converting molecular objects into SMILES strings.  Hosting these models on Hugging Face ensures efficient scaling and seamless updates without significant downtime. When a user uploads a reaction image file, the image is processed following the previously described Workflow. The outputs are aggregated and displayed in the web interface, allowing users to visualize parsed reaction data in both human-readable and machine-readable formats.

For efficient performance, the Gradio-based interface enables real-time interaction, providing a responsive user experience. The backend leverages the computational resources of Hugging Face Spaces to manage model inference tasks efficiently. This architecture supports fast parallel processing and minimizes latency during complex reaction image parsing. RxnIM.web is openly accessible at \url{https://huggingface.co/spaces/CYF200127/RxnIM}. The complete source code is available on GitHub at \url{https://github.com/CYF2000127/RxnIM}.

\section*{Data availability}
All data used for this article are available in \url{https://github.com/CYF2000127/RxnIM},
which contains the annotated datasets and test and train and validation split. Intermediate files for each reaction data parsing task reported in this method are stored in this repository with corresponding documentation. The model checkpoints can be downloaded directly from \url{https://huggingface.co/datasets/CYF200127/RxnIM}.
\section*{Code availability}
The code used for this article is available in \url{https://github.com/CYF2000127/RxnIM} alongside the data. The code includes Python scripts for synthetic dataset generation, preprocessing, model training, and model evaluation on the train and test sets presented in this publication. It also provides a gradio web demo to use the model directly.

\section*{Acknowledgements}
We thank the Information Technology Services Center (ITSC) in HKUST for providing the HPC3 and SuperPod Cluster as our computational resources.

\section*{Authors' contributions}
YC wrote the main manuscript and developed the model. 
YC and CTL prepared all figures and tables collaboratively. 
YC, HG, and CTL designed all experiments collaboratively.
HC, YH, JS, and HG supervised the work.
All authors contributed to the manuscript. All authors read and approved the final manuscript.

\section*{Ethics Declarations}
\paragraph{Competing interests}
The authors declare no competing financial interest.

\paragraph{Ethics approval and consent to participate}
Not applicable.

\paragraph{Consent for publication}
Not applicable.

\paragraph{Funding}
HKUST (Project No. R9251, Z1269)

\bibliographystyle{achemso}
\bibliography{bibliography}

\providecommand{\latin}[1]{#1}
\makeatletter
\providecommand{\doi}
  {\begingroup\let\do\@makeother\dospecials
  \catcode`\{=1 \catcode`\}=2 \doi@aux}
\providecommand{\doi@aux}[1]{\endgroup\texttt{#1}}
\makeatother
\providecommand*\mcitethebibliography{\thebibliography}
\csname @ifundefined\endcsname{endmcitethebibliography}  {\let\endmcitethebibliography\endthebibliography}{}
\begin{mcitethebibliography}{45}
\providecommand*\natexlab[1]{#1}
\providecommand*\mciteSetBstSublistMode[1]{}
\providecommand*\mciteSetBstMaxWidthForm[2]{}
\providecommand*\mciteBstWouldAddEndPuncttrue
  {\def\EndOfBibitem{\unskip.}}
\providecommand*\mciteBstWouldAddEndPunctfalse
  {\let\EndOfBibitem\relax}
\providecommand*\mciteSetBstMidEndSepPunct[3]{}
\providecommand*\mciteSetBstSublistLabelBeginEnd[3]{}
\providecommand*\EndOfBibitem{}
\mciteSetBstSublistMode{f}
\mciteSetBstMaxWidthForm{subitem}{(\alph{mcitesubitemcount})}
\mciteSetBstSublistLabelBeginEnd
  {\mcitemaxwidthsubitemform\space}
  {\relax}
  {\relax}

\bibitem[Staker \latin{et~al.}(2019)Staker, Marshall, Abel, and McQuaw]{1staker2019molecular}
Staker,~J.; Marshall,~K.; Abel,~R.; McQuaw,~C.~M. Molecular structure extraction from documents using deep learning. \emph{Journal of chemical information and modeling} \textbf{2019}, \emph{59}, 1017--1029\relax
\mciteBstWouldAddEndPuncttrue
\mciteSetBstMidEndSepPunct{\mcitedefaultmidpunct}
{\mcitedefaultendpunct}{\mcitedefaultseppunct}\relax
\EndOfBibitem
\bibitem[Beard and Cole(2020)Beard, and Cole]{2beard2020chemschematicresolver}
Beard,~E.~J.; Cole,~J.~M. ChemSchematicResolver: a toolkit to decode 2D chemical diagrams with labels and R-groups into annotated chemical named entities. \emph{Journal of chemical information and modeling} \textbf{2020}, \emph{60}, 2059--2072\relax
\mciteBstWouldAddEndPuncttrue
\mciteSetBstMidEndSepPunct{\mcitedefaultmidpunct}
{\mcitedefaultendpunct}{\mcitedefaultseppunct}\relax
\EndOfBibitem
\bibitem[Wilary and Cole(2021)Wilary, and Cole]{3wilary2021reactiondataextractor}
Wilary,~D.~M.; Cole,~J.~M. ReactionDataExtractor: a tool for automated extraction of information from chemical reaction schemes. \emph{Journal of Chemical Information and Modeling} \textbf{2021}, \emph{61}, 4962--4974\relax
\mciteBstWouldAddEndPuncttrue
\mciteSetBstMidEndSepPunct{\mcitedefaultmidpunct}
{\mcitedefaultendpunct}{\mcitedefaultseppunct}\relax
\EndOfBibitem
\bibitem[Wilary and Cole(2023)Wilary, and Cole]{4wilary2023reactiondataextractor}
Wilary,~D.~M.; Cole,~J.~M. ReactionDataExtractor 2.0: A deep learning approach for data extraction from chemical reaction schemes. \emph{Journal of Chemical Information and Modeling} \textbf{2023}, \emph{63}, 6053--6067\relax
\mciteBstWouldAddEndPuncttrue
\mciteSetBstMidEndSepPunct{\mcitedefaultmidpunct}
{\mcitedefaultendpunct}{\mcitedefaultseppunct}\relax
\EndOfBibitem
\bibitem[Qian \latin{et~al.}(2023)Qian, Guo, Tu, Coley, and Barzilay]{5qian2023rxnscribe}
Qian,~Y.; Guo,~J.; Tu,~Z.; Coley,~C.~W.; Barzilay,~R. RxnScribe: A sequence generation model for reaction diagram parsing. \emph{Journal of Chemical Information and Modeling} \textbf{2023}, \emph{63}, 4030--4041\relax
\mciteBstWouldAddEndPuncttrue
\mciteSetBstMidEndSepPunct{\mcitedefaultmidpunct}
{\mcitedefaultendpunct}{\mcitedefaultseppunct}\relax
\EndOfBibitem
\bibitem[Jessop \latin{et~al.}(2011)Jessop, Adams, Willighagen, Hawizy, and Murray-Rust]{11jessop2011oscar4}
Jessop,~D.~M.; Adams,~S.~E.; Willighagen,~E.~L.; Hawizy,~L.; Murray-Rust,~P. OSCAR4: a flexible architecture for chemical text-mining. \emph{Journal of cheminformatics} \textbf{2011}, \emph{3}, 41\relax
\mciteBstWouldAddEndPuncttrue
\mciteSetBstMidEndSepPunct{\mcitedefaultmidpunct}
{\mcitedefaultendpunct}{\mcitedefaultseppunct}\relax
\EndOfBibitem
\bibitem[Hawizy \latin{et~al.}(2011)Hawizy, Jessop, Adams, and Murray-Rust]{12hawizy2011chemicaltagger}
Hawizy,~L.; Jessop,~D.~M.; Adams,~N.; Murray-Rust,~P. ChemicalTagger: A tool for semantic text-mining in chemistry. \emph{Journal of cheminformatics} \textbf{2011}, \emph{3}, 1--13\relax
\mciteBstWouldAddEndPuncttrue
\mciteSetBstMidEndSepPunct{\mcitedefaultmidpunct}
{\mcitedefaultendpunct}{\mcitedefaultseppunct}\relax
\EndOfBibitem
\bibitem[Lowe(2012)]{13lowe2012extraction}
Lowe,~D.~M. Extraction of chemical structures and reactions from the literature. Ph.D.\ thesis, 2012\relax
\mciteBstWouldAddEndPuncttrue
\mciteSetBstMidEndSepPunct{\mcitedefaultmidpunct}
{\mcitedefaultendpunct}{\mcitedefaultseppunct}\relax
\EndOfBibitem
\bibitem[Swain and Cole(2016)Swain, and Cole]{14swain2016chemdataextractor}
Swain,~M.~C.; Cole,~J.~M. ChemDataExtractor: a toolkit for automated extraction of chemical information from the scientific literature. \emph{Journal of chemical information and modeling} \textbf{2016}, \emph{56}, 1894--1904\relax
\mciteBstWouldAddEndPuncttrue
\mciteSetBstMidEndSepPunct{\mcitedefaultmidpunct}
{\mcitedefaultendpunct}{\mcitedefaultseppunct}\relax
\EndOfBibitem
\bibitem[Steiner \latin{et~al.}(2019)Steiner, Wolf, Glatzel, Andreou, Granda, Keenan, Hinkley, Aragon-Camarasa, Kitson, Angelone, \latin{et~al.} others]{18steiner2019organic}
Steiner,~S.; Wolf,~J.; Glatzel,~S.; Andreou,~A.; Granda,~J.~M.; Keenan,~G.; Hinkley,~T.; Aragon-Camarasa,~G.; Kitson,~P.~J.; Angelone,~D.; others Organic synthesis in a modular robotic system driven by a chemical programming language. \emph{Science} \textbf{2019}, \emph{363}, eaav2211\relax
\mciteBstWouldAddEndPuncttrue
\mciteSetBstMidEndSepPunct{\mcitedefaultmidpunct}
{\mcitedefaultendpunct}{\mcitedefaultseppunct}\relax
\EndOfBibitem
\bibitem[Vaucher \latin{et~al.}(2020)Vaucher, Zipoli, Geluykens, Nair, Schwaller, and Laino]{19vaucher2020automated}
Vaucher,~A.~C.; Zipoli,~F.; Geluykens,~J.; Nair,~V.~H.; Schwaller,~P.; Laino,~T. Automated extraction of chemical synthesis actions from experimental procedures. \emph{Nature communications} \textbf{2020}, \emph{11}, 3601\relax
\mciteBstWouldAddEndPuncttrue
\mciteSetBstMidEndSepPunct{\mcitedefaultmidpunct}
{\mcitedefaultendpunct}{\mcitedefaultseppunct}\relax
\EndOfBibitem
\bibitem[Mayfield \latin{et~al.}(2018)Mayfield, Lowe, and Sayle]{15mayfield2018pistachio}
Mayfield,~J.; Lowe,~D.; Sayle,~R. Pistachio. \emph{Patent.[Online]. Available: https://www. nextmovesoftware. com/pistachio. html} \textbf{2018}, \relax
\mciteBstWouldAddEndPunctfalse
\mciteSetBstMidEndSepPunct{\mcitedefaultmidpunct}
{}{\mcitedefaultseppunct}\relax
\EndOfBibitem
\bibitem[Lowe and Mayfield(2020)Lowe, and Mayfield]{16lowe2020extraction}
Lowe,~D.~M.; Mayfield,~J. Extraction of Reactions from Patents using Grammars. CLEF (Working Notes). 2020\relax
\mciteBstWouldAddEndPuncttrue
\mciteSetBstMidEndSepPunct{\mcitedefaultmidpunct}
{\mcitedefaultendpunct}{\mcitedefaultseppunct}\relax
\EndOfBibitem
\bibitem[Nguyen \latin{et~al.}(2020)Nguyen, Zhai, Yoshikawa, Fang, Druckenbrodt, Thorne, Hoessel, Akhondi, Cohn, Baldwin, \latin{et~al.} others]{17nguyen2020chemu}
Nguyen,~D.~Q.; Zhai,~Z.; Yoshikawa,~H.; Fang,~B.; Druckenbrodt,~C.; Thorne,~C.; Hoessel,~R.; Akhondi,~S.~A.; Cohn,~T.; Baldwin,~T.; others ChEMU: named entity recognition and event extraction of chemical reactions from patents. Advances in Information Retrieval: 42nd European Conference on IR Research, ECIR 2020, Lisbon, Portugal, April 14--17, 2020, Proceedings, Part II 42. 2020; pp 572--579\relax
\mciteBstWouldAddEndPuncttrue
\mciteSetBstMidEndSepPunct{\mcitedefaultmidpunct}
{\mcitedefaultendpunct}{\mcitedefaultseppunct}\relax
\EndOfBibitem
\bibitem[Guo \latin{et~al.}(2021)Guo, Ibanez-Lopez, Gao, Quach, Coley, Jensen, and Barzilay]{20guo2021automated}
Guo,~J.; Ibanez-Lopez,~A.~S.; Gao,~H.; Quach,~V.; Coley,~C.~W.; Jensen,~K.~F.; Barzilay,~R. Automated chemical reaction extraction from scientific literature. \emph{Journal of chemical information and modeling} \textbf{2021}, \emph{62}, 2035--2045\relax
\mciteBstWouldAddEndPuncttrue
\mciteSetBstMidEndSepPunct{\mcitedefaultmidpunct}
{\mcitedefaultendpunct}{\mcitedefaultseppunct}\relax
\EndOfBibitem
\bibitem[Zhong \latin{et~al.}(2023)Zhong, Ouyang, Jiao, Kargupta, Luo, Shen, Zhou, Zhong, Liu, Li, \latin{et~al.} others]{21zhong2023reaction}
Zhong,~M.; Ouyang,~S.; Jiao,~Y.; Kargupta,~P.; Luo,~L.; Shen,~Y.; Zhou,~B.; Zhong,~X.; Liu,~X.; Li,~H.; others Reaction Miner: An Integrated System for Chemical Reaction Extraction from Textual Data. Proceedings of the 2023 Conference on Empirical Methods in Natural Language Processing: System Demonstrations. 2023; pp 389--402\relax
\mciteBstWouldAddEndPuncttrue
\mciteSetBstMidEndSepPunct{\mcitedefaultmidpunct}
{\mcitedefaultendpunct}{\mcitedefaultseppunct}\relax
\EndOfBibitem
\bibitem[Qian \latin{et~al.}(2023)Qian, Guo, Tu, Li, Coley, and Barzilay]{qian2023molscribe}
Qian,~Y.; Guo,~J.; Tu,~Z.; Li,~Z.; Coley,~C.~W.; Barzilay,~R. MolScribe: robust molecular structure recognition with image-to-graph generation. \emph{Journal of Chemical Information and Modeling} \textbf{2023}, \emph{63}, 1925--1934\relax
\mciteBstWouldAddEndPuncttrue
\mciteSetBstMidEndSepPunct{\mcitedefaultmidpunct}
{\mcitedefaultendpunct}{\mcitedefaultseppunct}\relax
\EndOfBibitem
\bibitem[Filippov and Nicklaus(2009)Filippov, and Nicklaus]{22filippov2009optical}
Filippov,~I.~V.; Nicklaus,~M.~C. Optical structure recognition software to recover chemical information: OSRA, an open source solution. 2009\relax
\mciteBstWouldAddEndPuncttrue
\mciteSetBstMidEndSepPunct{\mcitedefaultmidpunct}
{\mcitedefaultendpunct}{\mcitedefaultseppunct}\relax
\EndOfBibitem
\bibitem[Rajan \latin{et~al.}(2020)Rajan, Zielesny, and Steinbeck]{23rajan2020decimer}
Rajan,~K.; Zielesny,~A.; Steinbeck,~C. DECIMER: towards deep learning for chemical image recognition. \emph{Journal of Cheminformatics} \textbf{2020}, \emph{12}, 65\relax
\mciteBstWouldAddEndPuncttrue
\mciteSetBstMidEndSepPunct{\mcitedefaultmidpunct}
{\mcitedefaultendpunct}{\mcitedefaultseppunct}\relax
\EndOfBibitem
\bibitem[Oldenhof \latin{et~al.}(2020)Oldenhof, Arany, Moreau, and Simm]{24oldenhof2020chemgrapher}
Oldenhof,~M.; Arany,~A.; Moreau,~Y.; Simm,~J. ChemGrapher: optical graph recognition of chemical compounds by deep learning. \emph{Journal of chemical information and modeling} \textbf{2020}, \emph{60}, 4506--4517\relax
\mciteBstWouldAddEndPuncttrue
\mciteSetBstMidEndSepPunct{\mcitedefaultmidpunct}
{\mcitedefaultendpunct}{\mcitedefaultseppunct}\relax
\EndOfBibitem
\bibitem[Chen \latin{et~al.}(2021)Chen, Saxena, Li, Fleet, and Hinton]{chen2021pix2seq}
Chen,~T.; Saxena,~S.; Li,~L.; Fleet,~D.~J.; Hinton,~G. Pix2seq: A language modeling framework for object detection. \emph{arXiv preprint arXiv:2109.10852} \textbf{2021}, \relax
\mciteBstWouldAddEndPunctfalse
\mciteSetBstMidEndSepPunct{\mcitedefaultmidpunct}
{}{\mcitedefaultseppunct}\relax
\EndOfBibitem
\bibitem[OpenAI(2022)]{openai2022chatgpt}
OpenAI,~T. Chatgpt: Optimizing language models for dialogue. OpenAI. 2022\relax
\mciteBstWouldAddEndPuncttrue
\mciteSetBstMidEndSepPunct{\mcitedefaultmidpunct}
{\mcitedefaultendpunct}{\mcitedefaultseppunct}\relax
\EndOfBibitem
\bibitem[Brown \latin{et~al.}(2020)Brown, Mann, Ryder, Subbiah, Kaplan, Dhariwal, Neelakantan, Shyam, Sastry, Askell, \latin{et~al.} others]{31brown2020languageGPT3}
Brown,~T.; Mann,~B.; Ryder,~N.; Subbiah,~M.; Kaplan,~J.~D.; Dhariwal,~P.; Neelakantan,~A.; Shyam,~P.; Sastry,~G.; Askell,~A.; others Language models are few-shot learners. \emph{Advances in neural information processing systems} \textbf{2020}, \emph{33}, 1877--1901\relax
\mciteBstWouldAddEndPuncttrue
\mciteSetBstMidEndSepPunct{\mcitedefaultmidpunct}
{\mcitedefaultendpunct}{\mcitedefaultseppunct}\relax
\EndOfBibitem
\bibitem[Ouyang \latin{et~al.}(2022)Ouyang, Wu, Jiang, Almeida, Wainwright, Mishkin, Zhang, Agarwal, Slama, Ray, \latin{et~al.} others]{ouyang2022trainingINSGPT}
Ouyang,~L.; Wu,~J.; Jiang,~X.; Almeida,~D.; Wainwright,~C.; Mishkin,~P.; Zhang,~C.; Agarwal,~S.; Slama,~K.; Ray,~A.; others Training language models to follow instructions with human feedback. \emph{Advances in neural information processing systems} \textbf{2022}, \emph{35}, 27730--27744\relax
\mciteBstWouldAddEndPuncttrue
\mciteSetBstMidEndSepPunct{\mcitedefaultmidpunct}
{\mcitedefaultendpunct}{\mcitedefaultseppunct}\relax
\EndOfBibitem
\bibitem[Raffel \latin{et~al.}(2020)Raffel, Shazeer, Roberts, Lee, Narang, Matena, Zhou, Li, and Liu]{raffel2020exploringT5}
Raffel,~C.; Shazeer,~N.; Roberts,~A.; Lee,~K.; Narang,~S.; Matena,~M.; Zhou,~Y.; Li,~W.; Liu,~P.~J. Exploring the limits of transfer learning with a unified text-to-text transformer. \emph{Journal of machine learning research} \textbf{2020}, \emph{21}, 1--67\relax
\mciteBstWouldAddEndPuncttrue
\mciteSetBstMidEndSepPunct{\mcitedefaultmidpunct}
{\mcitedefaultendpunct}{\mcitedefaultseppunct}\relax
\EndOfBibitem
\bibitem[Chowdhery \latin{et~al.}(2023)Chowdhery, Narang, Devlin, Bosma, Mishra, Roberts, Barham, Chung, Sutton, Gehrmann, \latin{et~al.} others]{chowdhery2023palm}
Chowdhery,~A.; Narang,~S.; Devlin,~J.; Bosma,~M.; Mishra,~G.; Roberts,~A.; Barham,~P.; Chung,~H.~W.; Sutton,~C.; Gehrmann,~S.; others Palm: Scaling language modeling with pathways. \emph{Journal of Machine Learning Research} \textbf{2023}, \emph{24}, 1--113\relax
\mciteBstWouldAddEndPuncttrue
\mciteSetBstMidEndSepPunct{\mcitedefaultmidpunct}
{\mcitedefaultendpunct}{\mcitedefaultseppunct}\relax
\EndOfBibitem
\bibitem[Zhang \latin{et~al.}(2022)Zhang, Roller, Goyal, Artetxe, Chen, Chen, Dewan, Diab, Li, Lin, \latin{et~al.} others]{zhang2022opt}
Zhang,~S.; Roller,~S.; Goyal,~N.; Artetxe,~M.; Chen,~M.; Chen,~S.; Dewan,~C.; Diab,~M.; Li,~X.; Lin,~X.~V.; others Opt: Open pre-trained transformer language models. \emph{arXiv preprint arXiv:2205.01068} \textbf{2022}, \relax
\mciteBstWouldAddEndPunctfalse
\mciteSetBstMidEndSepPunct{\mcitedefaultmidpunct}
{}{\mcitedefaultseppunct}\relax
\EndOfBibitem
\bibitem[Alayrac \latin{et~al.}(2022)Alayrac, Donahue, Luc, Miech, Barr, Hasson, Lenc, Mensch, Millican, Reynolds, \latin{et~al.} others]{alayrac2022flamingo}
Alayrac,~J.-B.; Donahue,~J.; Luc,~P.; Miech,~A.; Barr,~I.; Hasson,~Y.; Lenc,~K.; Mensch,~A.; Millican,~K.; Reynolds,~M.; others Flamingo: a visual language model for few-shot learning. \emph{Advances in neural information processing systems} \textbf{2022}, \emph{35}, 23716--23736\relax
\mciteBstWouldAddEndPuncttrue
\mciteSetBstMidEndSepPunct{\mcitedefaultmidpunct}
{\mcitedefaultendpunct}{\mcitedefaultseppunct}\relax
\EndOfBibitem
\bibitem[Shen \latin{et~al.}(2024)Shen, Song, Tan, Li, Lu, and Zhuang]{shen2024hugginggpt}
Shen,~Y.; Song,~K.; Tan,~X.; Li,~D.; Lu,~W.; Zhuang,~Y. Hugginggpt: Solving ai tasks with chatgpt and its friends in hugging face. \emph{Advances in Neural Information Processing Systems} \textbf{2024}, \emph{36}\relax
\mciteBstWouldAddEndPuncttrue
\mciteSetBstMidEndSepPunct{\mcitedefaultmidpunct}
{\mcitedefaultendpunct}{\mcitedefaultseppunct}\relax
\EndOfBibitem
\bibitem[Yang \latin{et~al.}(2023)Yang, Li, Wang, Lin, Azarnasab, Ahmed, Liu, Liu, Zeng, and Wang]{yang2023mmGPT}
Yang,~Z.; Li,~L.; Wang,~J.; Lin,~K.; Azarnasab,~E.; Ahmed,~F.; Liu,~Z.; Liu,~C.; Zeng,~M.; Wang,~L. Mm-react: Prompting chatgpt for multimodal reasoning and action. \emph{arXiv preprint arXiv:2303.11381} \textbf{2023}, \relax
\mciteBstWouldAddEndPunctfalse
\mciteSetBstMidEndSepPunct{\mcitedefaultmidpunct}
{}{\mcitedefaultseppunct}\relax
\EndOfBibitem
\bibitem[Wu \latin{et~al.}(2023)Wu, Yin, Qi, Wang, Tang, and Duan]{wu2023visualGPT}
Wu,~C.; Yin,~S.; Qi,~W.; Wang,~X.; Tang,~Z.; Duan,~N. Visual chatgpt: Talking, drawing and editing with visual foundation models. \emph{arXiv preprint arXiv:2303.04671} \textbf{2023}, \relax
\mciteBstWouldAddEndPunctfalse
\mciteSetBstMidEndSepPunct{\mcitedefaultmidpunct}
{}{\mcitedefaultseppunct}\relax
\EndOfBibitem
\bibitem[Li \latin{et~al.}(2023)Li, He, Wang, Li, Wang, Luo, Wang, Wang, and Qiao]{li2023videochat}
Li,~K.; He,~Y.; Wang,~Y.; Li,~Y.; Wang,~W.; Luo,~P.; Wang,~Y.; Wang,~L.; Qiao,~Y. Videochat: Chat-centric video understanding. \emph{arXiv preprint arXiv:2305.06355} \textbf{2023}, \relax
\mciteBstWouldAddEndPunctfalse
\mciteSetBstMidEndSepPunct{\mcitedefaultmidpunct}
{}{\mcitedefaultseppunct}\relax
\EndOfBibitem
\bibitem[Liu \latin{et~al.}(2023)Liu, He, Wang, Wang, Wang, Chen, Zhang, Yang, Li, Yu, \latin{et~al.} others]{liu2023internchat}
Liu,~Z.; He,~Y.; Wang,~W.; Wang,~W.; Wang,~Y.; Chen,~S.; Zhang,~Q.; Yang,~Y.; Li,~Q.; Yu,~J.; others Internchat: Solving vision-centric tasks by interacting with chatbots beyond language. \emph{arXiv preprint arXiv:2305.05662} \textbf{2023}, \relax
\mciteBstWouldAddEndPunctfalse
\mciteSetBstMidEndSepPunct{\mcitedefaultmidpunct}
{}{\mcitedefaultseppunct}\relax
\EndOfBibitem
\bibitem[Liu \latin{et~al.}(2023)Liu, Li, Li, Yu, Huang, Peng, Liu, Chen, Li, Jin, \latin{et~al.} others]{liu2023hidden}
Liu,~Y.; Li,~Z.; Li,~H.; Yu,~W.; Huang,~M.; Peng,~D.; Liu,~M.; Chen,~M.; Li,~C.; Jin,~L.; others On the hidden mystery of ocr in large multimodal models. \emph{arXiv preprint arXiv:2305.07895} \textbf{2023}, \relax
\mciteBstWouldAddEndPunctfalse
\mciteSetBstMidEndSepPunct{\mcitedefaultmidpunct}
{}{\mcitedefaultseppunct}\relax
\EndOfBibitem
\bibitem[Martori(2022)]{OChemR}
Martori,~D.,~M.;~Probst Machine Learning approach for chemical reactions digitalisation. \emph{https://github.com/markmartorilopez/OChemR} \textbf{2022}, \relax
\mciteBstWouldAddEndPunctfalse
\mciteSetBstMidEndSepPunct{\mcitedefaultmidpunct}
{}{\mcitedefaultseppunct}\relax
\EndOfBibitem
\bibitem[Wang \latin{et~al.}(2024)Wang, Chen, Chen, Wu, Zhu, Zeng, Luo, Lu, Zhou, Qiao, \latin{et~al.} others]{wang2024visionllm}
Wang,~W.; Chen,~Z.; Chen,~X.; Wu,~J.; Zhu,~X.; Zeng,~G.; Luo,~P.; Lu,~T.; Zhou,~J.; Qiao,~Y.; others Visionllm: Large language model is also an open-ended decoder for vision-centric tasks. \emph{Advances in Neural Information Processing Systems} \textbf{2024}, \emph{36}\relax
\mciteBstWouldAddEndPuncttrue
\mciteSetBstMidEndSepPunct{\mcitedefaultmidpunct}
{\mcitedefaultendpunct}{\mcitedefaultseppunct}\relax
\EndOfBibitem
\bibitem[He \latin{et~al.}(2016)He, Zhang, Ren, and Sun]{he2016deep}
He,~K.; Zhang,~X.; Ren,~S.; Sun,~J. Deep residual learning for image recognition. Proceedings of the IEEE conference on computer vision and pattern recognition. 2016; pp 770--778\relax
\mciteBstWouldAddEndPuncttrue
\mciteSetBstMidEndSepPunct{\mcitedefaultmidpunct}
{\mcitedefaultendpunct}{\mcitedefaultseppunct}\relax
\EndOfBibitem
\bibitem[Devlin \latin{et~al.}(2018)Devlin, Chang, Lee, and Toutanova]{devlin2018bert}
Devlin,~J.; Chang,~M.-W.; Lee,~K.; Toutanova,~K. Bert: Pre-training of deep bidirectional transformers for language understanding. \emph{arXiv preprint arXiv:1810.04805} \textbf{2018}, \relax
\mciteBstWouldAddEndPunctfalse
\mciteSetBstMidEndSepPunct{\mcitedefaultmidpunct}
{}{\mcitedefaultseppunct}\relax
\EndOfBibitem
\bibitem[Vaswani \latin{et~al.}(2017)Vaswani, Shazeer, Parmar, Uszkoreit, Jones, Gomez, Kaiser, and Polosukhin]{vaswani2017attention}
Vaswani,~A.; Shazeer,~N.; Parmar,~N.; Uszkoreit,~J.; Jones,~L.; Gomez,~A.~N.; Kaiser,~{\L}.; Polosukhin,~I. Attention is all you need. \emph{Advances in neural information processing systems} \textbf{2017}, \emph{30}\relax
\mciteBstWouldAddEndPuncttrue
\mciteSetBstMidEndSepPunct{\mcitedefaultmidpunct}
{\mcitedefaultendpunct}{\mcitedefaultseppunct}\relax
\EndOfBibitem
\bibitem[Chen \latin{et~al.}(2023)Chen, Zhang, Zeng, Zhang, Zhu, and Zhao]{chen2023shikra}
Chen,~K.; Zhang,~Z.; Zeng,~W.; Zhang,~R.; Zhu,~F.; Zhao,~R. Shikra: Unleashing Multimodal LLM's Referential Dialogue Magic. \emph{arXiv preprint arXiv:2306.15195} \textbf{2023}, \relax
\mciteBstWouldAddEndPunctfalse
\mciteSetBstMidEndSepPunct{\mcitedefaultmidpunct}
{}{\mcitedefaultseppunct}\relax
\EndOfBibitem
\bibitem[Zhang \latin{et~al.}(2024)Zhang, Sun, Chen, Xiao, Shao, Zhang, Liu, Chen, and Luo]{zhang2024gpt4roi}
Zhang,~S.; Sun,~P.; Chen,~S.; Xiao,~M.; Shao,~W.; Zhang,~W.; Liu,~Y.; Chen,~K.; Luo,~P. GPT4RoI: Instruction Tuning Large Language Model on Region-of-Interest. 2024\relax
\mciteBstWouldAddEndPuncttrue
\mciteSetBstMidEndSepPunct{\mcitedefaultmidpunct}
{\mcitedefaultendpunct}{\mcitedefaultseppunct}\relax
\EndOfBibitem
\bibitem[Peng \latin{et~al.}(2023)Peng, Wang, Dong, Hao, Huang, Ma, and Wei]{peng2023kosmos}
Peng,~Z.; Wang,~W.; Dong,~L.; Hao,~Y.; Huang,~S.; Ma,~S.; Wei,~F. Kosmos-2: Grounding multimodal large language models to the world. \emph{arXiv preprint arXiv:2306.14824} \textbf{2023}, \relax
\mciteBstWouldAddEndPunctfalse
\mciteSetBstMidEndSepPunct{\mcitedefaultmidpunct}
{}{\mcitedefaultseppunct}\relax
\EndOfBibitem
\bibitem[Zhu \latin{et~al.}(2020)Zhu, Su, Lu, Li, Wang, and Dai]{zhu2020deformable}
Zhu,~X.; Su,~W.; Lu,~L.; Li,~B.; Wang,~X.; Dai,~J. Deformable detr: Deformable transformers for end-to-end object detection. \emph{arXiv preprint arXiv:2010.04159} \textbf{2020}, \relax
\mciteBstWouldAddEndPunctfalse
\mciteSetBstMidEndSepPunct{\mcitedefaultmidpunct}
{}{\mcitedefaultseppunct}\relax
\EndOfBibitem
\bibitem[Touvron \latin{et~al.}(2023)Touvron, Lavril, Izacard, Martinet, Lachaux, Lacroix, Rozi{\`e}re, Goyal, Hambro, Azhar, \latin{et~al.} others]{touvron2023llama}
Touvron,~H.; Lavril,~T.; Izacard,~G.; Martinet,~X.; Lachaux,~M.-A.; Lacroix,~T.; Rozi{\`e}re,~B.; Goyal,~N.; Hambro,~E.; Azhar,~F.; others Llama: Open and efficient foundation language models. \emph{arXiv preprint arXiv:2302.13971} \textbf{2023}, \relax
\mciteBstWouldAddEndPunctfalse
\mciteSetBstMidEndSepPunct{\mcitedefaultmidpunct}
{}{\mcitedefaultseppunct}\relax
\EndOfBibitem
\bibitem[Touvron \latin{et~al.}(2023)Touvron, Martin, Stone, Albert, Almahairi, Babaei, Bashlykov, Batra, Bhargava, Bhosale, \latin{et~al.} others]{touvron2023llama2}
Touvron,~H.; Martin,~L.; Stone,~K.; Albert,~P.; Almahairi,~A.; Babaei,~Y.; Bashlykov,~N.; Batra,~S.; Bhargava,~P.; Bhosale,~S.; others Llama 2: Open foundation and fine-tuned chat models. \emph{arXiv preprint arXiv:2307.09288} \textbf{2023}, \relax
\mciteBstWouldAddEndPunctfalse
\mciteSetBstMidEndSepPunct{\mcitedefaultmidpunct}
{}{\mcitedefaultseppunct}\relax
\EndOfBibitem
\end{mcitethebibliography}

% \clearpage
% \appendix

% \section{Experimental procedures}
% \subsection{Insect repellent}

\clearpage

\end{sloppypar}
\end{document}